\documentclass[conference]{IEEEtran}
\usepackage{times}

\usepackage[numbers]{natbib}
\usepackage{multicol}
\usepackage[bookmarks=true]{hyperref}

\usepackage{amsmath}
\usepackage{amsthm}
\usepackage{amsfonts}
\usepackage{bm}
\usepackage{algorithm}
\usepackage{algpseudocode}
\usepackage{subcaption}
\usepackage{todonotes}
\usepackage{verbatim}

\usepackage{environ}
\NewEnviron{problem}[1]{%
\begin{center}\fbox{\parbox{3in}{%
    {\centering\scshape #1\par}%
    \parskip=1ex
    \everypar{\hangindent=1em}%
    \BODY
}}\end{center}}

\usepackage{etex,etoolbox}
\usepackage{amsthm,amssymb}
\usepackage{thmtools, thm-restate}
\usepackage{environ}

\makeatletter
\providecommand{\@fourthoffour}[4]{#4}
\newcommand\fixstatement[2][\proofname\space of]{%
  \ifcsname thmt@original@#2\endcsname
    \AtEndEnvironment{#2}{%
      \xdef\pat@label{\expandafter\expandafter\expandafter
        \@fourthoffour\csname thmt@original@#2\endcsname\space\@currentlabel}%
      \xdef\pat@proofof{\@nameuse{pat@proofof@#2}}%
    }%
  \else
    \AtEndEnvironment{#2}{%
      \xdef\pat@label{\expandafter\expandafter\expandafter
        \@fourthoffour\csname #1\endcsname\space\@currentlabel}%
      \xdef\pat@proofof{\@nameuse{pat@proofof@#2}}%
    }%
  \fi
  \@namedef{pat@proofof@#2}{#1}%
}

\globtoksblk\prooftoks{1000}
\newcounter{proofcount}

\NewEnviron{proofatend}{%
  \edef\next{%
    \noexpand\begin{proof}[\pat@proofof\space\pat@label]%
    \unexpanded\expandafter{\BODY}}%
  \global\toks\numexpr\prooftoks+\value{proofcount}\relax=\expandafter{\next\end{proof}}
  \stepcounter{proofcount}
}

\def\printproofs{%
  \count@=\z@
  \loop
    \the\toks\numexpr\prooftoks+\count@\relax
    \ifnum\count@<\value{proofcount}%
    \advance\count@\@ne
  \repeat}
\makeatother

\newtheorem{thm}{Theorem}
\newtheorem{defn}{Definition}
\newtheorem{lemma}{Lemma}

\fixstatement{thm}
\fixstatement{lemma}

\begin{document}
\title{Provably Safe Robot Navigation with Obstacle Uncertainty}

% You will get a Paper-ID when submitting a pdf file to the conference system
\author{Brian Axelrod, Leslie Pack Kaelbling and Tom\'as Lozano-P\'erez}

%\author{\authorblockN{Michael Shell}
%\authorblockA{School of Electrical and\\Computer Engineering\\
%Georgia Institute of Technology\\
%Atlanta, Georgia 30332--0250\\
%Email: mshell@ece.gatech.edu}
%\and
%\authorblockA{Twentieth Century Fox\\
%\and
%\authorblockN{James Kirk\\ and Montgomery Scott}
%\authorblockA{Starfleet Academy\\
%San Francisco, California 96678-2391\\

% avoiding spaces at the end of the author lines is not a problem with
% conference papers because we don't use \thanks or \IEEEmembership

%Homer Simpson\authorrefmark{2},
%Montgomery Scott\authorrefmark{3} and
%Eldon Tyrell\authorrefmark{4}}
%Telephone: (800) 555--1212, Fax: (888) 555--1212}
%\authorblockA{\authorrefmark{4}Tyrell Inc., 123 Replicant Street, Los Angeles, California 90210--4321}}

\maketitle

\begin{abstract}
  As drones and autonomous cars become more widespread it is becoming
  increasingly important that robots can operate safely under
  realistic conditions. The noisy information fed into real systems
  means that robots must use estimates of the environment to plan
  navigation. Efficiently guaranteeing that the resulting motion plans
  are safe under these circumstances has proved difficult. We examine
  how to guarantee that a trajectory or policy is safe with only
  imperfect observations of the environment. We examine the
  implications of various mathematical formalisms of safety and arrive
  at a mathematical notion of safety of a long-term execution, even
  when conditioned on observational information.  We present efficient
  algorithms that can prove that trajectories or policies are safe
  with much tighter bounds than in previous work. Notably, the
  complexity of the environment does not affect our method's ability
  to evaluate if a trajectory or policy is safe. We then use these
  safety checking methods to design a safe variant of the RRT planning
  algorithm. 
\end{abstract}

\IEEEpeerreviewmaketitle

\section{Introduction}
\subsection{Motivation}
Safe and reliable operation of a robot in a cluttered environment can
be difficult to achieve due to noisy and partial observations of the
state of both the world and the robot. As autonomous systems leave the
factory floor and become more pervasive in the form of drones and
self-driving cars, it is becoming increasingly important to understand
how to design systems that will not fail under these real-world
conditions. While it is important that these systems be safe, it is
also important they do not operate so conservatively as to be
ineffective. They must have a strong understanding of when they take
risks so they can avoid them, but still operate efficiently.   

While most previous work focuses on robot state uncertainty, this
paper focuses on safe navigation when the locations and geometries of
these obstacles are uncertain. We focus on algorithms that find safety
``certificates"---easily verifiable proofs that the trajectory or
policy is safe. We examine two implications of the algorithms. First,
the computational complexity of reasoning about uncertainty can be
quite low. Second, the mathematics surrounding robot safety can have
surprising behavior. We demonstrate how these tools can be used to
design a motion planner guaranteed to give only safe plans, and inform
the design of more general systems that make decisions under
uncertainty. 

\begin{figure}
\centering
\includegraphics[width=\columnwidth]{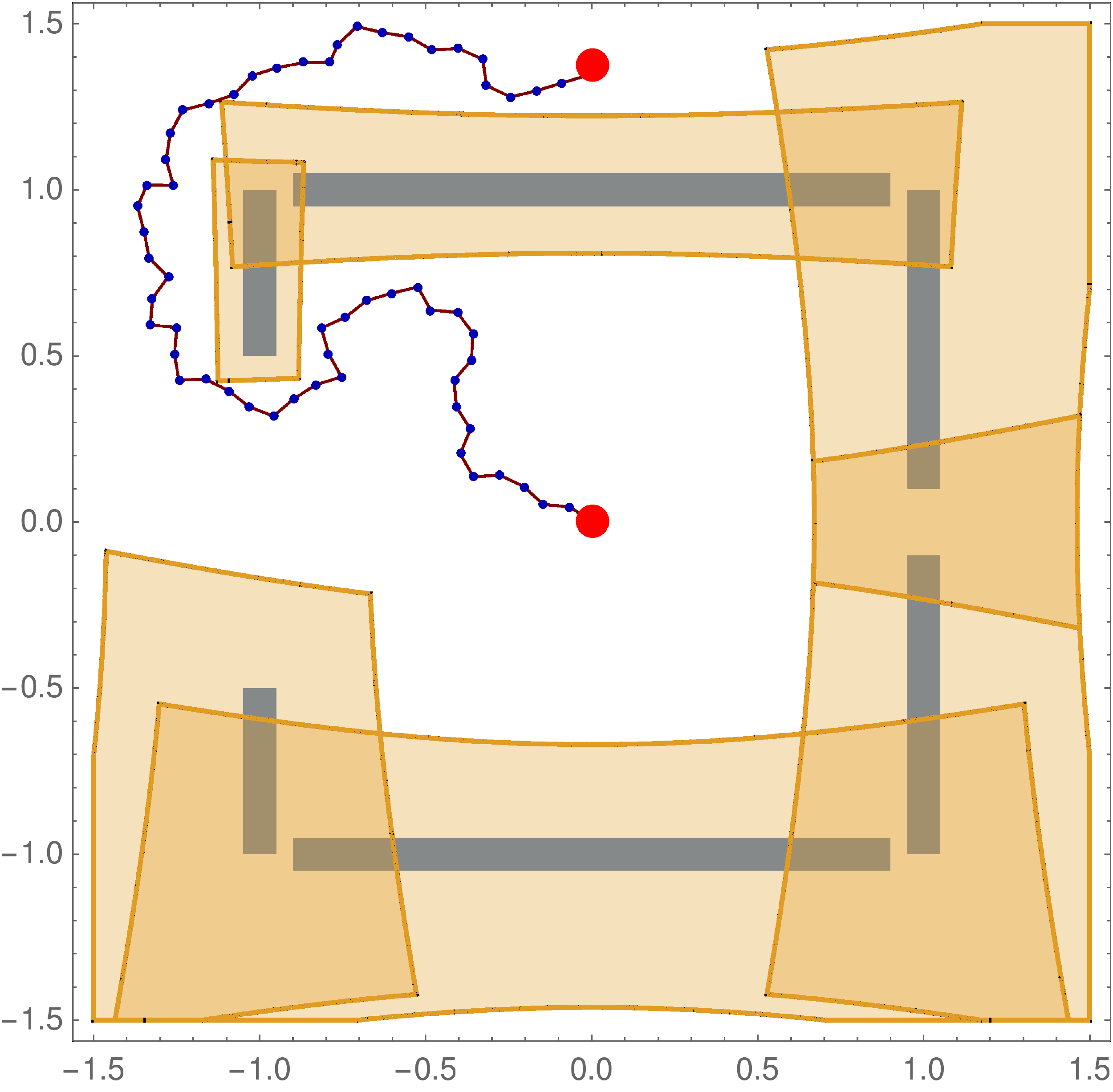}
\caption{The desired trajectory found by the planner shown with its specialized shadows that certify the probability of collision as less than $0.26\%$.}
\label{fig:safetraj}
\end{figure}

\subsection{Problem Formulation}
We consider two settings.  In the {\em off-line} setting we have a
fixed set of information about the environment and are searching for
an open-loop trajectory. In the {\em on-line} setting the robot has
access to a stream of observations and can change its trajectory as a
function of new information; the problem is to find a policy, a
function from observations to actions, that allow the robot to adapt
to changing circumstances.  We show that different notions of safety
are required for the two cases to ensure that the robot can guarantee
a low probability of collision throughout its entire execution.

Safety in the offline setting amounts to staying out of regions likely
to be contained within obstacles, and can be analyzed by computing
geometric bounds on obstacles for which we have only partial
information. Safety in the online setting builds on offline safety by
requiring that the robot respect a contract with respect to the
aggregate lifetime risk of operation while always having a guaranteed
safe trajectory available to it.

We develop a general framework for analyzing safety and provide an
example of applying this framework to a specific model of random
geometry. We wish to emphasize that this framework can be applied to a
wide variety of models beyond the example shown here.  

Our framework operates in generality in $\mathbb R^n$ and assumes that
obstacles are polytopes in $\mathbb R^n$. In this paper we focus on
examples of typical robotic domains in $\mathbb R^2$ and $\mathbb
R^3$. We ensure safety by verifying that the swept volume of a robot's
trajectory is unlikely to collide with any obstacles. These swept
volumes can be computed geometrically, or, for dynamical systems, via
SOS programs \cite{majumdar2012algebraic}.  
%\lpk{But, maybe you have a broader set of cases in mind in which case you need to spell out two or three of them more clearly.}
%\baxelrod{Hmmm, mostly I don't want to restrict myself. The practical application is $n=2,3$ though the formulas I derive all work for arbitrary spaces. You could have a obstacles moving with time etc, if you wanted to be really fancy forex. Not sure how to talk about possible extensions though.}

We say that a trajectory, a map from time to robot configurations
${\mathcal Q}$, $\tau: [0,\infty) \rightarrow \mathcal Q$ is
$\epsilon-$safe if the swept volume of the robot along trajectory $\tau$
intersects an obstacle with probability less than
$\epsilon$. Formally, if $A$ is the event that the swept volume of
$\tau$ intersects an obstacle, then $\tau$ is \textit{$\epsilon-$safe}
in the offline sense if $P(A \mid \tau) \le \epsilon$.

We say that a policy, a map from observation history $O$, state history $H$
and time to a trajectory $\tau$, $\pi: O \times H \times
[0,\infty)\rightarrow \tau$ is $\epsilon-$safe if,  under all sequences of
observations, $P(A \mid \pi) \le \epsilon$. This notion of safety will
be referred to as \textit{policy safety}; it is a departure from
previous models of robot safety, capturing
the notion of a contract that the total risk over the
lifetime of the system always be less than $\epsilon$. 
\label{sec:policySafety}

The requirement that the safety condition hold under all observations
sequences is more conservative than the natural definition of
$P(A \mid \pi )$.  It crucial to prevent undesirable behavior that can
``cheat'' the definition of safety; this is discussed in detail in
section~\ref{sec:onlinesafety}.

\subsection{Related Work}
Planning under uncertainty has been studied extensively. Some
approaches operate in generality and compute complete policies
\cite{kaelbling1998planning} while operate online, computing a
plausible open-loop plan and updating it as more information is
acquired \cite{platt2010belief}.  

Generating plans that provide formal non-collision (safety) guarantees
when the environment is uncertain has proven difficult. Many methods
use heuristic strategies to try to ensure that the plans they generate are
unlikely to collide. One way of ensuring that a trajectory is safe is
simply staying sufficiently far away from obstacles. If the robot's
pose is uncertain this can be achieved by planning with a robot whose
shape is grown by an uncertainty
bound~\cite{bry2011rapidly}. Alternatively, if the 
obstacle geometry is uncertain, the area around estimated obstacles
can be expanded into a shadow whose size depends on the magnitude of
the uncertainty~\cite{kaelbling2013integrated,lee2013sigma}. %While earlier
%work by anonymous authors was able to provide formal guarantees, 
%this necessitated very conservative behavior~\cite{mypaper}. 

Another approach focuses on evaluating the probability that a
trajectory will collide. Monte-Carlo methods can evaluate the
probability of collision by sampling, but can be computationally
expensive when the likelihood of failure is very
small~\cite{montecarloLJESMP2015}. When the uncertainty is restricted
to Gaussian uncertainty on the robot's pose, probabilistic collision
checking can yield notable performance improvements~\cite{sun2016safe}\cite{Park2016EfficientPC}\cite{Park2016FastAB}.

Another perspective is finding a plan that is safe by construction. If
the system is modeled as a Markov Decision Process, formal
verification methods can be used to construct a plan that is guaranteed
to be safe~\cite{ding2013strategic}\cite{feyzabadi2016multi}. Recent 
work on methods that are based on signal temporal logic (STL) model have also 
uncertainty in obstacle geometry. With PrSTL \citet{prstl} explicitly
model uncertainty in the environment to help generate safe plans but
offer weaker guarantees than our work. 

\subsection{Contributions}
This paper makes three contributions. 
the first is a formal definition of online safety that provides risk
bounds on the entire execution of a policy.

The second contribution is an algorithm for efficiently verifying
offline safety with respect to polytopes with Gaussian distributed
faces (PGDFs) that is then generalized to the online case. In
comparison to previous methods, the quality of the resulting bound is
not dependent on the number of obstacles in the environment. The
presented algorithms produce a certificate, which allows another
system to efficiently verify that the actions about to be taken are
safe. For a maximal collision probability of $\epsilon$, the 
runtime of the algorithm grows as $\log \frac 1 \epsilon$ making
 it efficient even for very small $\epsilon$'s.

The third contribution is a modification to the RRT algorithm that
generates safe plans.  For any fixed $\epsilon$, the resulting planner
is guaranteed to only return trajectories for which the probability of
failure is less than $\epsilon$. We note that for $n$ obstacles, the
runtime of the RRT is increased only by a $\log n \log \frac 1
\epsilon$ factor, which suggests that reasoning about uncertainty can
come at a low computational cost.  A result of running this algorithm
is shown in figure~\ref{fig:safetraj}.

\section{Model for Random Geometry} 

Previous work on planning under uncertainty has often relied on the
notion of a shadow~\cite{kaelbling2013integrated, lee2013sigma}, which
is a volume that represents an uncertain estimate of the pose of an
object and the space that it may occupy.  A proper shadow is likely to
contain the true object; and even if the exact location of the object
is not known, it is sufficient to avoid the obstacle's shadow in order
to avoid the true obstacle. A shadow is essentially the geometric
equivalent of a confidence interval.

In order to provide strong guarantees about the safety of robot
trajectories we first formalize this notion of a shadow.  
\begin{defn}\label{def:shadow} A set $X$ is said to be an {\em
    $\epsilon-$shadow} of a random obstacle $O$ if $P( O \subset X)
  \ge 1 - \epsilon$.
\end{defn}
To be able to generate shadows with desired properties, we need to 
place some restrictions on the class of distributions from which
obstacles are drawn.

One way to arrive at a distribution on the shape and position of
obstacles is to imagine that sensor data is obtained in the form of
point clouds in which the points are segmented according to the face of
the obstacle to which they belong. Then, the points belonging to a
particular obstacle face can be used to perform a Bayesian linear
regression on the parameters of the
plane containing the face; given a Gaussian prior on the face-plane
parameters and under reasonable assumptions about the noise-generation
process, the posterior on the face-plane parameters will also be
Gaussian~\cite{bishop2006pattern, rasmussen2006gaussian}.

Recalling that a polytope $X$ is the intersection of halfspaces, we
can define a distribution over the parameters of polytopes with a
fixed number of faces. For $\bm x$
represented in homogeneous coordinates, a polytope $X$ can be
represented as 
$$X = \bigcap \bm n_i ^T \bm x \le 0\;\;.$$
When these normal vectors $\bm n_i$ are drawn from a Gaussian
distribution $\mathcal N \left ( \bm \mu_i, \bm
  \Sigma_i \right ) $, we will call this a polytope with
Gaussian distributed Faces (PGDF) with parameters $\bm \mu_i, \bm
\Sigma_i$. We do not assume that the normal vectors are drawn
independently and show that an independence assumption would yield
little additional tightness to our bounds.  

Using the PGDF model, we will be able to identify shadow regions that
are guaranteed to contain the obstacle with a probability greater than
$1 - \epsilon$ in most cases of interest.  There are degenerate
combinations of values of $\epsilon$, $\mu$, and $\Sigma$ for which
there is no well-defined shadow (consider the case in which the means
of all the face-planes go through the origin, for example).  In
addition, there are some cases in which the bounds we use for
constructing regions are not tight and so although a shadow region
exists, we cannot construct it.  Details of these cases are discussed
in the proofs in the supplementary material.

If we consider a single face, theorem~\ref{lemma:lgshape} identifies a
shadow region likely to contain the corresponding half-space. 

 \begin{thm}\label{lemma:lgshape} Consider $\epsilon \in (0,1)$,
   $\bm n \sim \mathcal N \left  (\bm \mu, \bm \Sigma \right)$ such
   that the combination of $\epsilon, \bm \mu, \bm \Sigma$ is
   non-degenerate. There exists a shadow $S$ s.t. $\{ \bm x \mid \bm n^T \bm x \ge 0 \}$ is
   contained within $S$ with probability at least $1 -
   \epsilon$. \end{thm}  

\begin{figure}[tp]
\centering
\includegraphics[width=0.4\columnwidth]{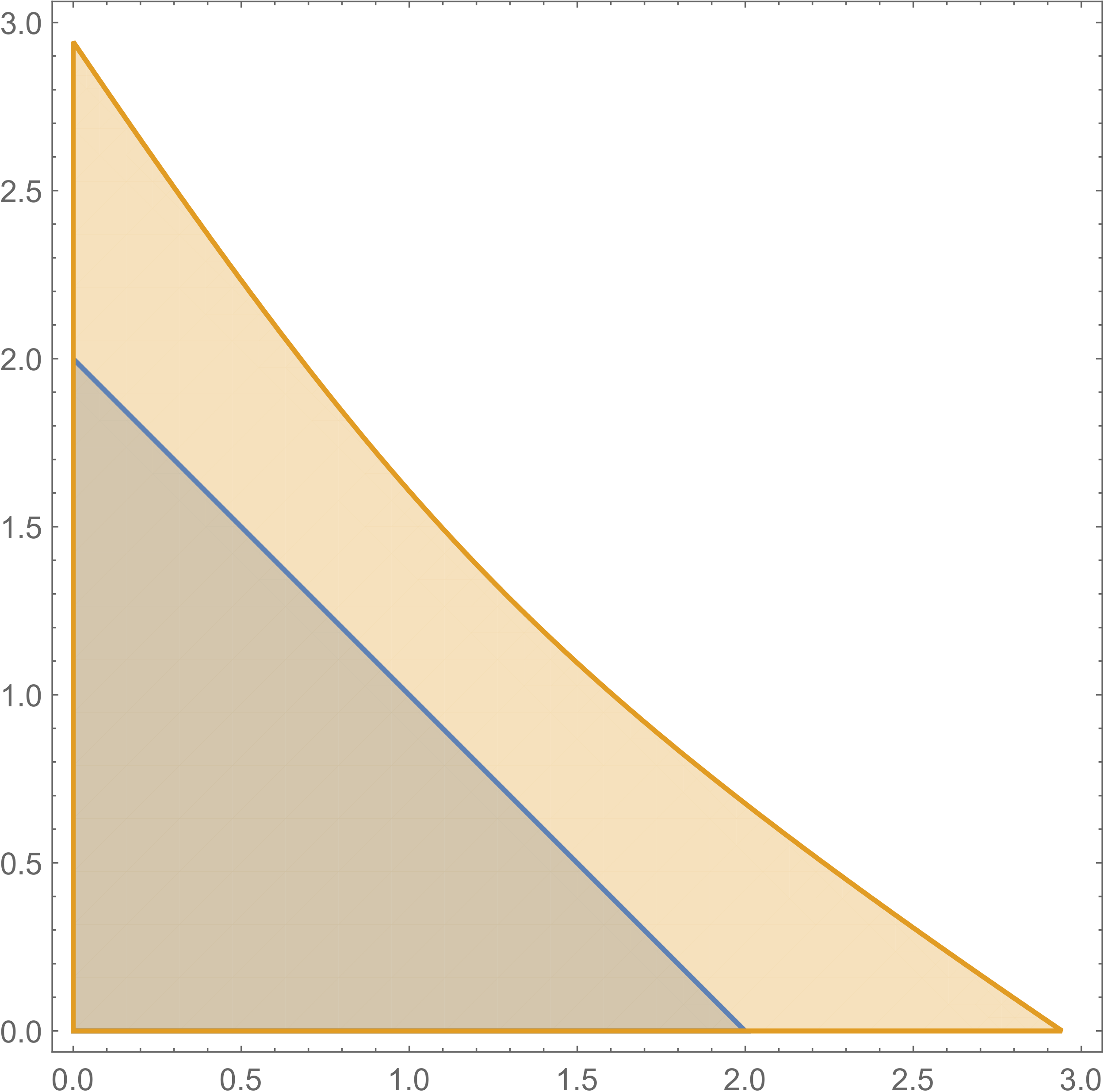}
\caption{An example of the region defined by theorem \ref{lemma:lgshape}. The darker region is the "mean" obstacle and the orange region contains 95\% of the obstacles generated by these parameters.}
\label{fig:lgshape}
\end{figure}

 While a detailed constructive proof is deferred to the supplemental
 materials, we present a sketch here. First, we identify a
 sufficiently high probability-mass region of half-spaces $\bm n$,
 which, by construction corresponds to a linear cone $C$ in the space
 of half-space parameters. We then take the set of $\bm x$'s in
 homogeneous coordinates that these half-spaces contain. The set of
 points not contained by any half-space is the polar cone of
 $C$. Converting back to non-homogeneous coordinates yields conic
 sections as seen in figure~\ref{fig:lgshape}.

In lemma \ref{lemma:polybuild} we generalize this notion to
polytopes. For an obstacle with $m$ faces, we can take the
intersection of the resulting $\frac \epsilon m$-shadows. A union bound
guarantees that the probability of any face not being contained in its
corresponding region is less than $\epsilon$. Thus the probability
that the polytope is not contained in this region is less than
$\epsilon$. In other words, lemma \ref{lemma:polybuild} constructs 
an $\epsilon$ shadow.

\begin{lemma}\label{lemma:polybuild} Consider a polytope defined by
  $\bigcap\limits_i \bm n^T \bm x \le 0$. Let $X_i$ be a set that
  contains the halfspace defined by $\bm n_i$ with probability at
  least $ 1- \epsilon_i$ (for example as in theorem
  \ref{lemma:lgshape}). Then $\bigcap\limits_i X_i$ contains the
  polytope with probability at least $1 - \sum\limits_i \epsilon_i$.  
\end{lemma}
\begin{comment}
    \begin{proofatend}
We note that $\bigcap\limits_i X_i$ contains the polytope if every $X_i$ contains its corresponding halfspace. Since the probability that $X_i$ does not contain its corresponding halfspace is $\epsilon_i$, a union bound gives us that the probability that any $X_i$ does not contain its corresponding halfspace is bounded by $\sum\limits_i \epsilon_i$. $\bigcap\limits_i X_i$ containing the polytope is the complement of this event, thus the probability that $\bigcap\limits_i X_i$ contains the polytope is at least $ 1 - \sum\limits_i \epsilon_o$
\end{proofatend}
\end{comment}
\subsection{Computing Obstacle Shadows}

\begin{figure}[tp]
\centering
\begin{tikzpicture}
  \node[anchor=south west,inner sep=0] at (0,0) {\includegraphics[width=0.3\textwidth]{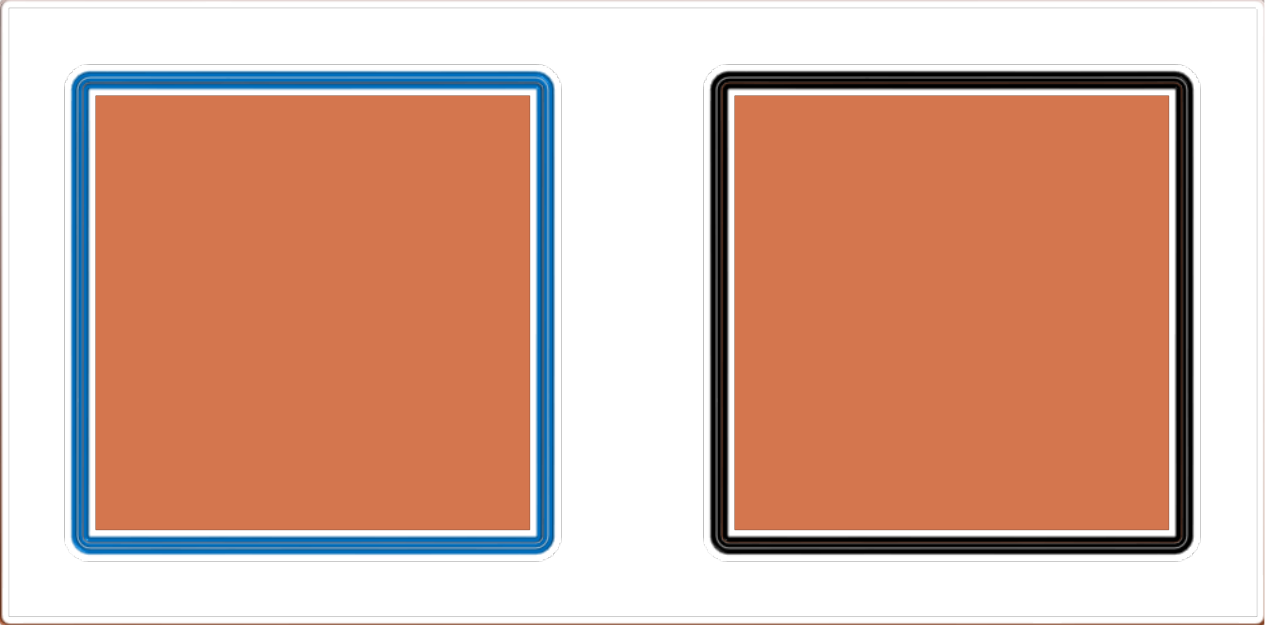}};
  \node[draw,align=left] at (1.3,3) {$P(A) = 0.5$};
  \node[draw,align=left] at (4.1,3) {$P(A^C) = 0.5$};
  \end{tikzpicture}
\caption{Both the blue and black squares are valid $0.5$-shadows, while the union of the two yellow areas is the set of points with probability at least $0.5$ of being in the square.}
\label{fig:shadowquirk1}
\end{figure}

\begin{figure}[tp]
\centering
\includegraphics[width=0.7\columnwidth]{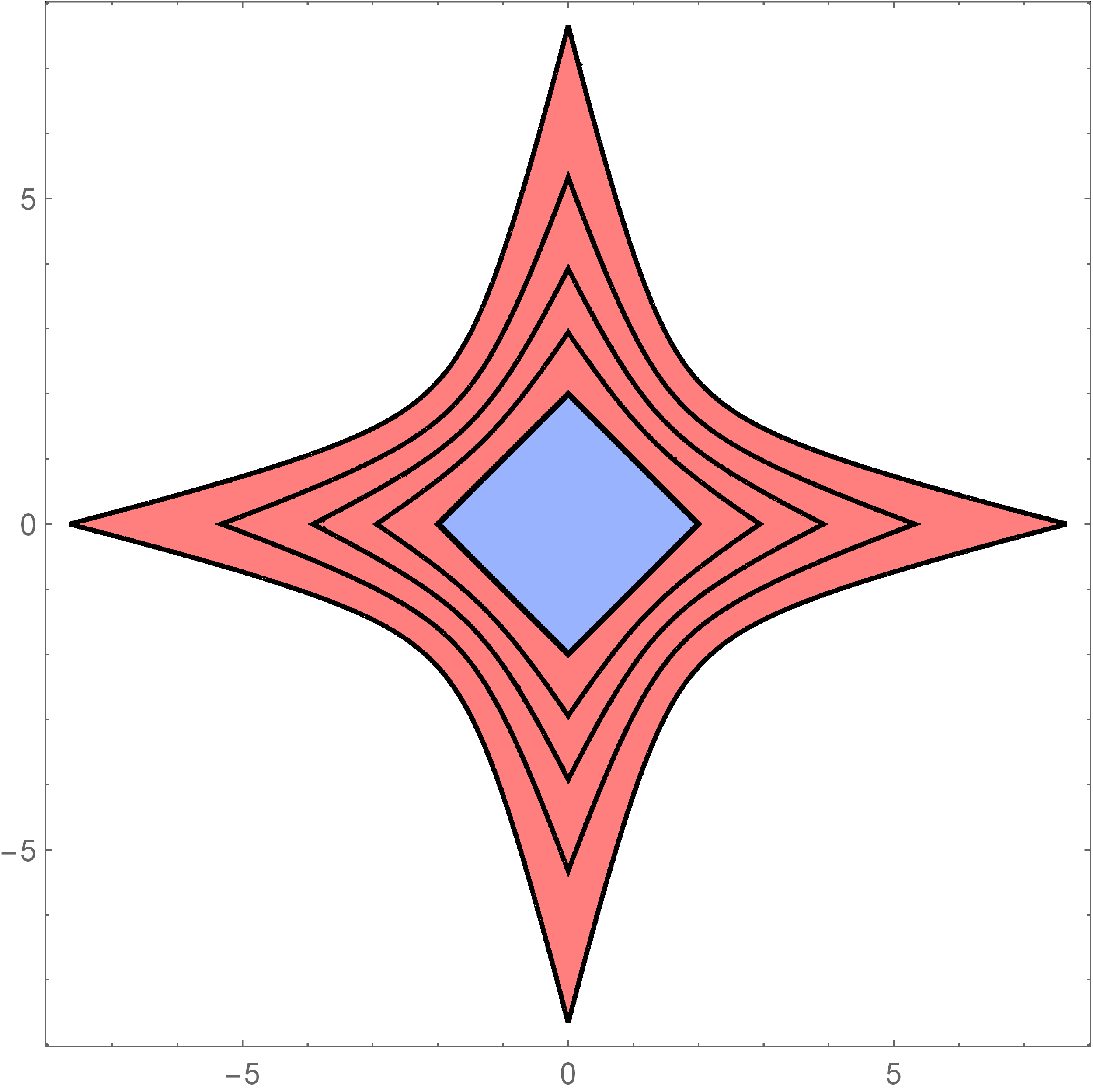}
\caption{The blue square represents the ``mean" estimated obstacle. Each outline in the red set represents a different probability shadow of the obstacle.}
\label{fig:manyshadows}
\end{figure}

We will use theorem~\ref{lemma:lgshape} and lemma~\ref{lemma:polybuild} 
to construct regions which we can prove are shadows. 
Before we continue, however, we note that obstacle shadows need not be
unique, or correspond to the set of points with probability greater
than $\epsilon$ of being inside the obstacle. Consider the case where
a fair coin flip determines the location of a square as shown in
figure \ref{fig:shadowquirk1}.
While there have been previous attempts to derive unsafe regions for
normally-distributed faces~\cite{prstl}, this lemma is stronger in
that it constructs a set that is likely contain the obstacle as
opposed to identifying points which are individually likely to be
inside the obstacle.

Recall that a PGDF obstacle is a random polytope with $m$ sides,
that is, the intersection of $m$ halfspaces. We can provide a shadow for
each halfspace using theorem \ref{lemma:lgshape}, and use lemma
\ref{lemma:polybuild} to combine their intersection into a shadow for
the estimated obstacle. The result is a $\epsilon-$shadow for a PGDF
obstacle. An sequence of such increasingly tight $\epsilon-$shadows is
shown in figure \ref{fig:manyshadows}. 

\begin{lemma}\label{lemma:myshadow} If an obstacle is PGDF with
  nondegenerate parameters and $m$ sides, we can construct an
  $\epsilon$-shadow as the intersection of the $\frac \epsilon m$ shadows of
  each of its sides. 
\end{lemma}
\begin{comment}
\begin{proofatend}
Let $X_i$ be the shadow constructed by Lemma 1 with parameter
$\epsilon/m$ for side $i$. Then the shadow $X = \bigcap\limits_i X_i$
contains the PGDF with probability at least $1 -
\sum\limits_i\epsilon_i/m = 1 - \epsilon$. 
\end{proofatend}
\end{comment}

\subsection{Shadows as Safety Certificates}

Above we showed that for a given $\epsilon$, we can easily compute a
shadow for an obstacle under our model. Since a shadow is likely to
contain the real object, the non-intersection of an $\epsilon-$shadow
with the swept volume of a trajectory guarantees that the probability
of colliding with the given obstacle is less than $\epsilon$. 
Theorem 
\ref{lemma:multiobstacle} generalizes this notion to multiple
obstacles using a union bound.  

\begin{thm}\label{lemma:multiobstacle} 
Let $X$ be the volume of 
  space that the robot may visit during its lifetime. If for a given set of
  obstacles, indexed by $i$, and their corresponding $\epsilon_i$
  shadows, $X$ does not intersect any shadow, then the probability of
  collision with any obstacle over the lifetime of the system is at
  most $\sum \epsilon_i$.
\end{thm}
{\bf Proof:} Please see the supplementary material.

\begin{comment}
\begin{proofatend}
Let $A_i$ be the event that the intersection of $X$ and obstacle $i$ is nonempty. Since the shadow of obstacle $i$ did not intersect $X$, this means that obstacle $i$ must not be contained by its shadow. The probability of this is less than $\epsilon_i$ by definition of $\epsilon-$shadow. Thus the probability of $A_i$ must be at most $\epsilon_i$. 

Applying a union bound gives us that $P(\bigcup A_i) \le \sum \epsilon_i$.
\end{proofatend}
\end{comment}

It is important that theorem~\ref{lemma:multiobstacle} does not depend
on anything but the intersection of the swept volume with the
obstacle shadows; it is independent of the trajectory's length and of
how close to the shadow it comes. The guarantees of the theorem hold
regardless of what the robot chooses to do in the space not occupied
by a shadow. 

Computing a shadow requires solving a relatively simple system of
equations;  then given a set of shadows and their corresponding
$\epsilon$'s, a collision check between the visited states and the
shadows is sufficient to verify that the proposed trajectory is
safe. This implies that safety is easy to verify computationally,
potentially enabling redundant safety checks without much
computational power. Alternatively, a secondary system can verify that
the robot is fulfilling its safety contract.  

One potential concern is the tightness of
theorem~\eqref{lemma:multiobstacle}. If the union bound is loose, it may
force 
us to fail to certify trajectories that are safe in practice. The
union bound used in the proof of theorem~\eqref{lemma:multiobstacle}
assumes the worst case correlation. If we assume that the system is to
certify that collisions are rare events, and the events are
independent, the union bound here ends up being almost tight.  

\begin{lemma} \label{lemma:uniontight} 
Given $n$ obstacles and their shadows, if
\begin{itemize}
\item the events that each obstacle is not contained in its
  shadow are independent, 
\item the probability that obstacles are
  not contained in their shadows is less than $\epsilon$, and
\item $\epsilon = O(\frac{\sqrt \delta}{n})$
\end{itemize}
then the difference between the true
probability of a shadow not containing the object, and the union
bound in theorem~\ref{lemma:multiobstacle} is less than $\delta$.   
\end{lemma}  
{\bf Proof:} Please see the supplementary material.

\begin{comment}
\begin{proofatend}
We note that the union bound is only loose when at least two events
happen at the same time, so it is sufficient to bound the probability
that two shadows fail to contain their corresponding PGDF. 

The probability of this happening for two given obstacles is at most $\epsilon^2$, and there are $n \choose 2$ such combinations. Thus by a union bound, the probability of two shadows failing to contain their obstacle is at most $n \choose 2 \epsilon^2$. If $\epsilon \le k \frac{\sqrt \delta}{n}$ for $k \ge \frac{1}{\sqrt 2}$, then the probability of this happening is at most $\delta$. 

Since the probability of more than one event happening at the same time is bounded by $\delta$, the probability given by the union bound is within $\delta$ of the true probability. 
\end{proofatend}
\end{comment}
%\lpk{this whole thing can be worded better.}
Another way of interpreting lemma \ref{lemma:uniontight} is that if the probability of failure is small, the union bound is close to tight. 

\begin{figure}[tp]
\centering
\includegraphics[width=0.5\columnwidth]{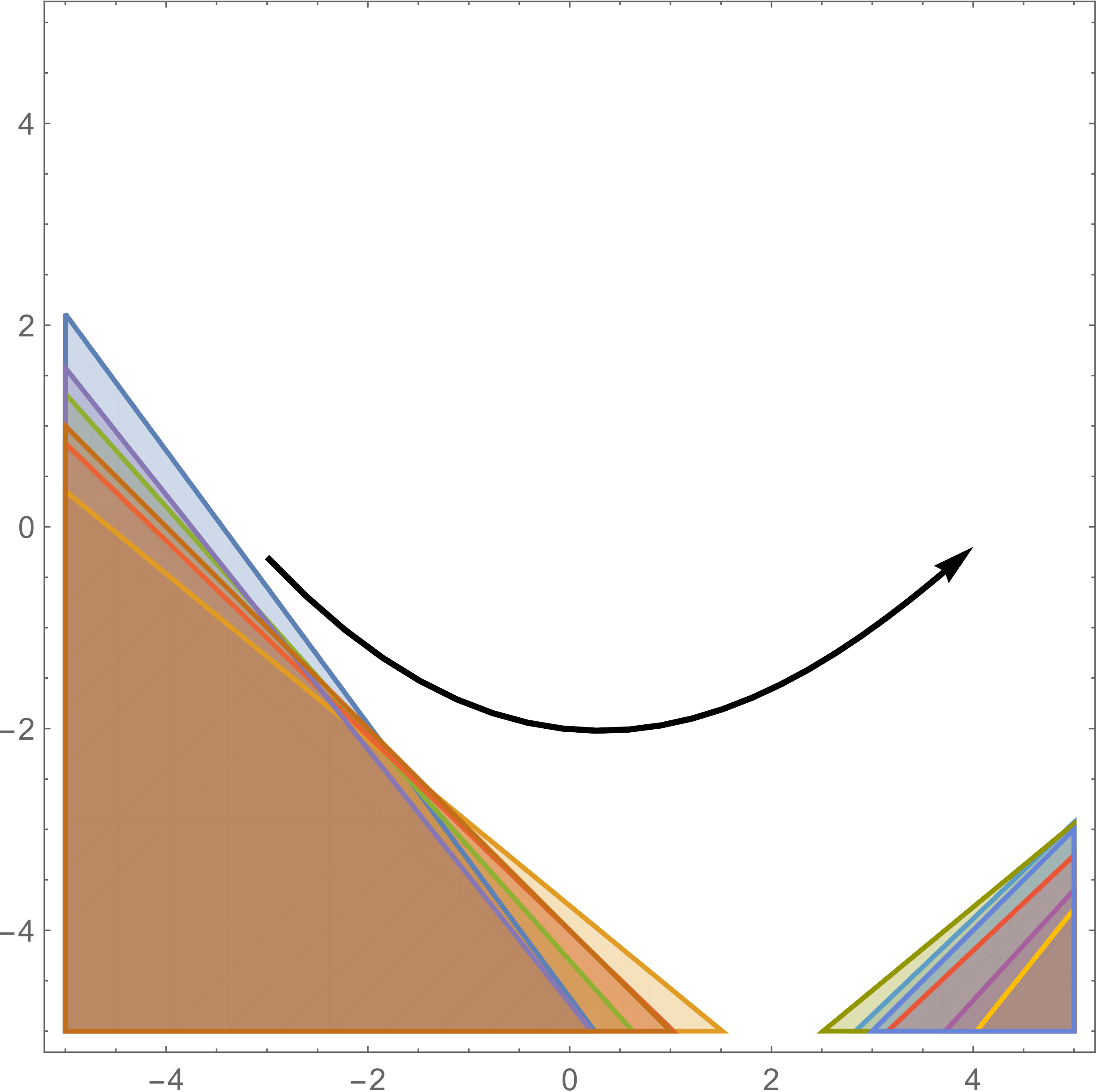}
\caption{This figure overlays several possible instances of the real obstacles given a single observation. The trajectory is much more likely to collide with a draw of the bottom left obstacle than the bottom right one. The two obstacles do not affect the safety of the trajectory equally. }
\label{fig:farawaysafe}
\end{figure}

Theorem \ref{lemma:multiobstacle} highlights a problem with systems that
do not search for a set of optimal shadows. If we allocate equal
$\epsilon$'s for every obstacle, then we are forced to pay as much for
an object far away and irrelevant to the trajectory as we are for
obstacles close to the trajectory. Figure~\ref{fig:farawaysafe}
suggests that it is optimal to have large shadows with very small
$\epsilon$'s for such irrelevant obstacles, and larger $\epsilon$'s
for obstacles that may present significant risk. Doing so requires
finding a good $\epsilon$-shadow pair for every obstacle. We present
an algorithm that finds this optimal certificate in the next section. 

\section{Algorithm For Finding Optimal Shadows}  
We will verify that trajectories are safe by finding a set of shadows
that proves the swept volume of the trajectory is unlikely to collide with an
obstacle. 
% The trajectory safety problem is formalized in algorithm
% \ref{alg:trajsafetydef}. 
% \begin{algorithm}
%  \caption{Trajectory Safety Verification}
%  \begin{algorithmic}[1]
%  \label{alg:trajsafetydef}
%  \renewcommand{\algorithmicrequire}{\textbf{Input:}}
%  \renewcommand{\algorithmicensure}{\textbf{Output:}}
%  \Require \begin{tabular}{p{7.5cm}}
%  $\epsilon$ - Maximum probability of failure                                                                              \\
%  $\epsilon_p$ - Maximum numerical error per shadow                                                                         \\
% $\{\bm \mu_{ij}, \bm \Sigma_{ij}\}$ - Parameters of estimated obstacles. Obstacles indexed by $i$, faces indexed by $j$. \\
% $X$ - the set of states which the system may visit. This may simply be the projection of the trajectory.                                 
% \end{tabular}
%  \Ensure  Return true if a certificate shows that the probability that $X$ intersects an obstacle is less than $\epsilon$ 
%  \end{algorithmic} 
%  \end{algorithm}
In order to minimize the number of scenarios in which a trajectory is
actually safe, but our system fails to certify it as such, we will search for
the optimal set of shadows for the given trajectory, allowing shadows
for distant obstacles to be larger than those for obstacles near the trajectory.
In order to understand the search for shadows of multiple obstacles we
first examine the case of a single obstacle.  

\subsection{Single Obstacle}
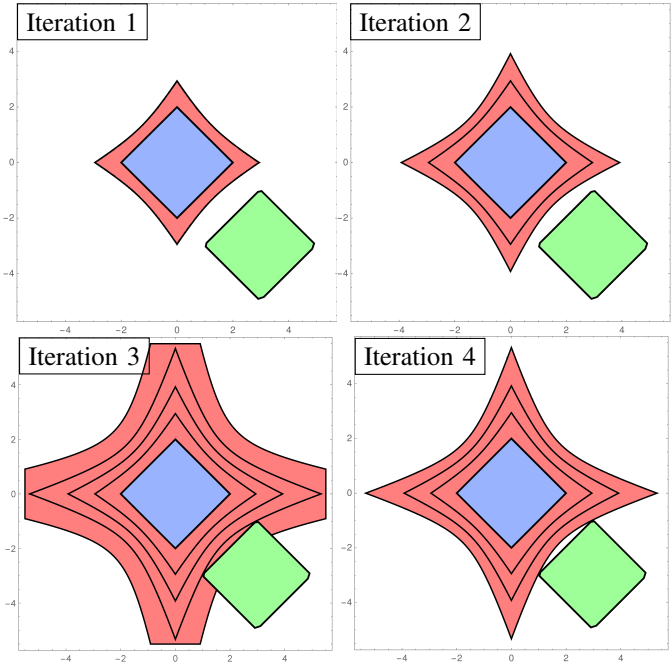
\begin{figure}[tp]
\centering
\begin{subfigure}{0.5\columnwidth}
  \centering
  \begin{tikzpicture}
  \node[anchor=south west,inner sep=0] at (0,0) {\includegraphics[width=\textwidth]{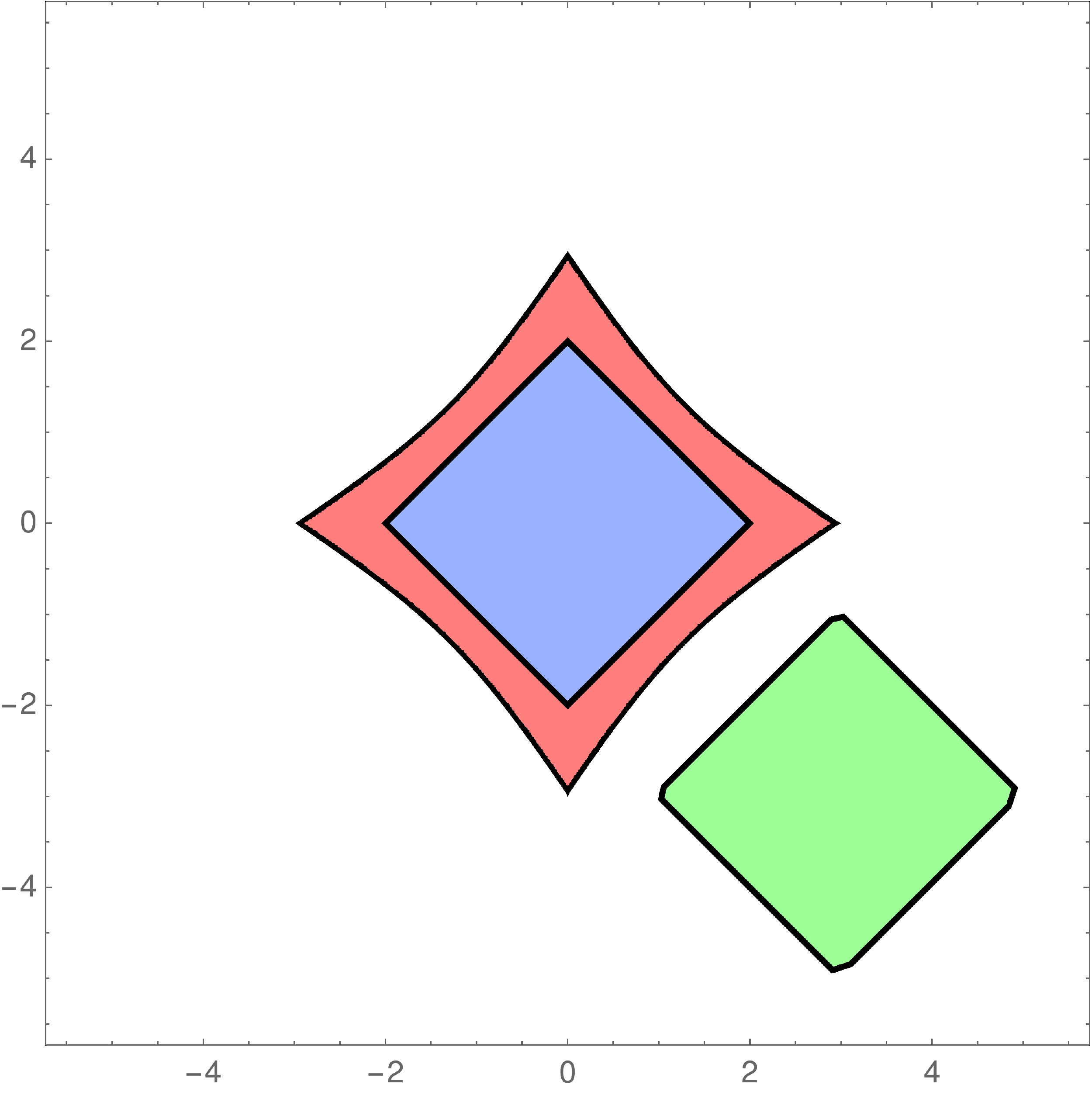}};
  \node[draw,align=left] at (1.05,4.2) {Iteration 1};
  \end{tikzpicture}
\end{subfigure}%
\begin{subfigure}{0.5\columnwidth}
  \centering
  \begin{tikzpicture}
  \node[anchor=south west,inner sep=0] at (0,0) {\includegraphics[width=\textwidth]{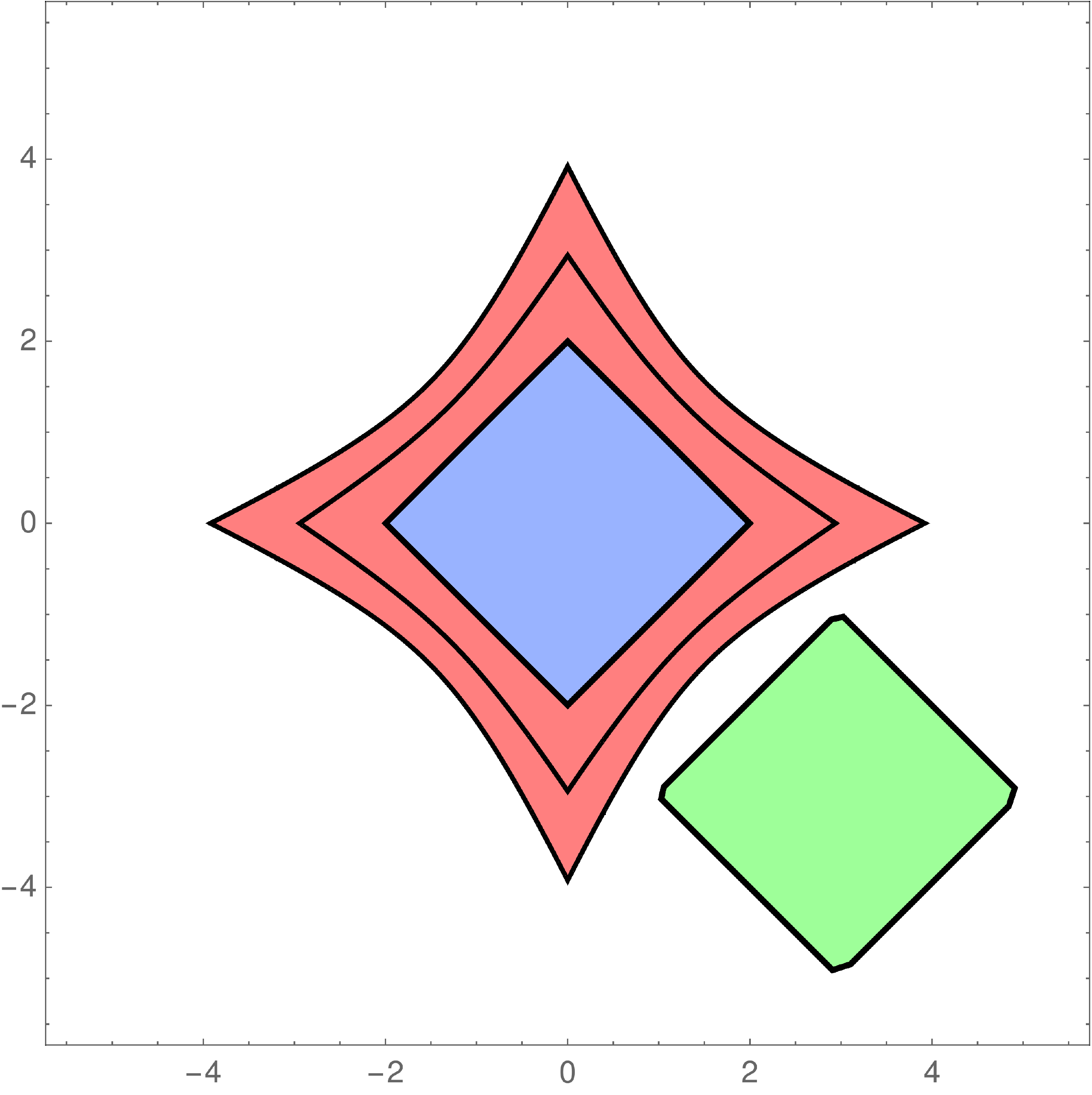}};
  \node[draw,align=left] at (1.05,4.2) {Iteration 2};
  \end{tikzpicture}
\end{subfigure}
\begin{subfigure}{0.49\columnwidth}
  \centering
  \begin{tikzpicture}
  \node[anchor=south west,inner sep=0] at (0,0) {\includegraphics[width=\textwidth]{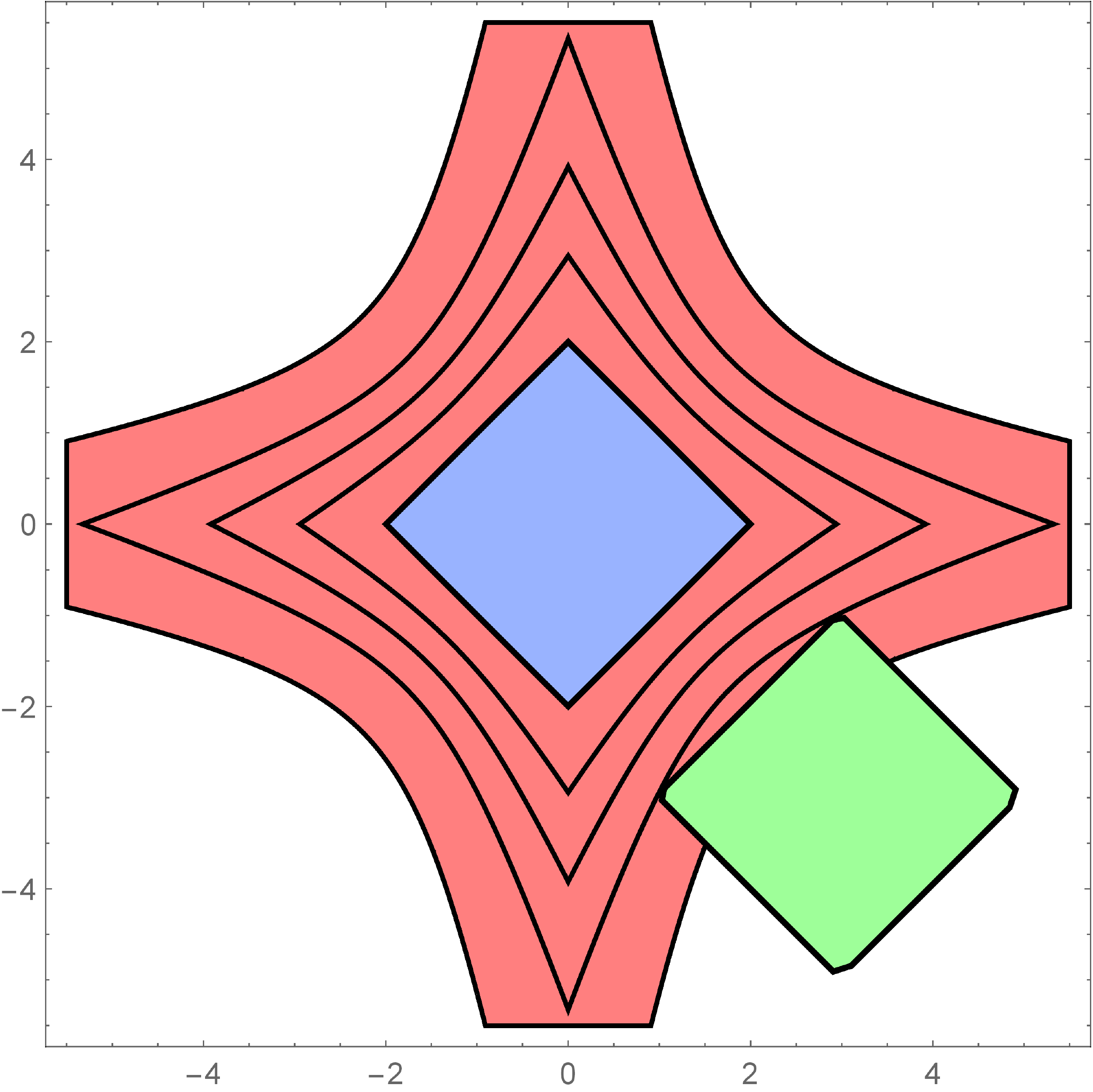}};
  \node[draw,align=left] at (1.05,4.1) {Iteration 3};
  \end{tikzpicture}
\end{subfigure} % 
\begin{subfigure}{0.49\columnwidth}
  \centering
  \begin{tikzpicture}
  \node[anchor=south west,inner sep=0] at (0,0) {\includegraphics[width=\textwidth]{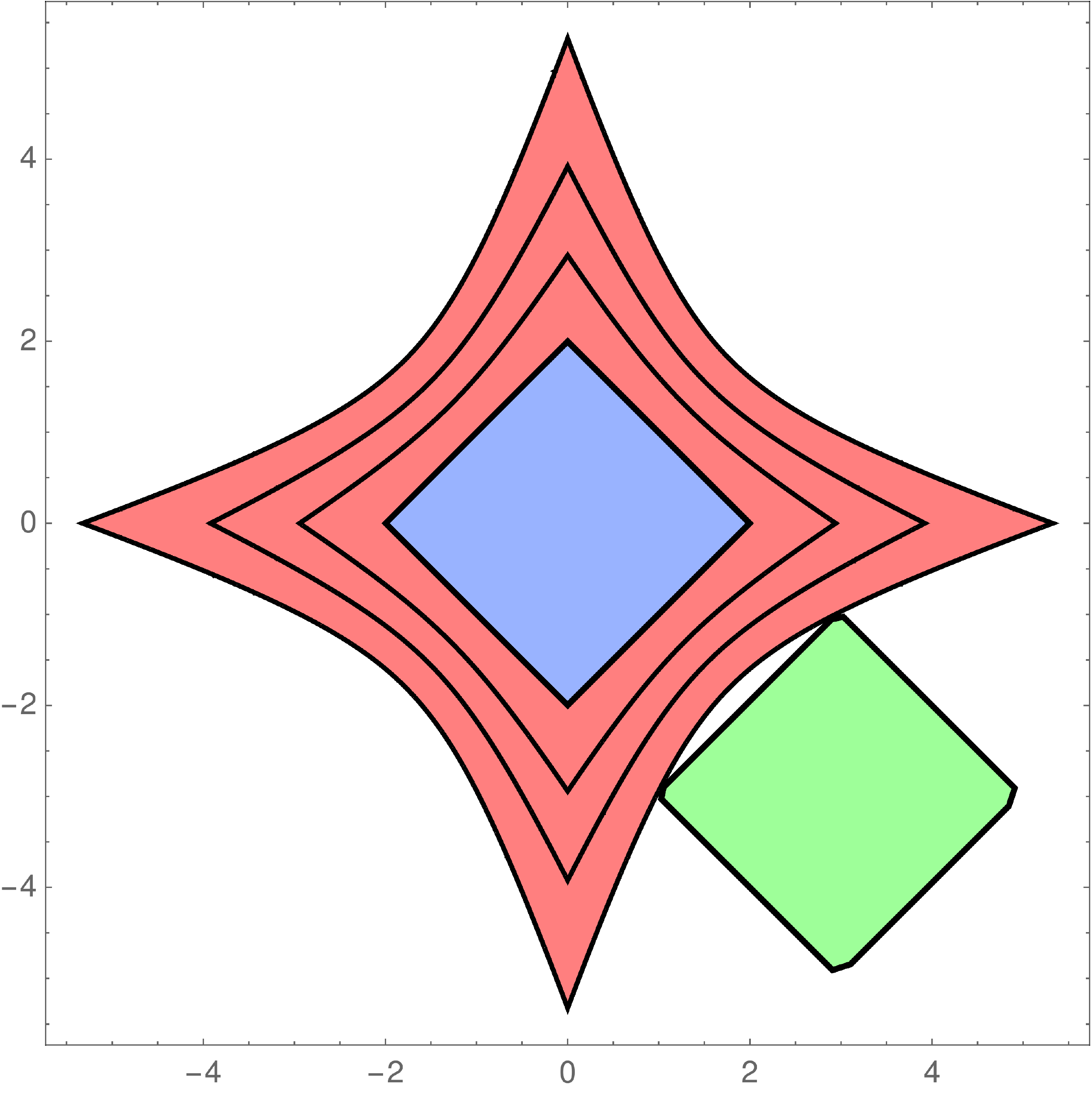}};
  \node[draw,align=left] at (1.05,4.1) {Iteration 4};
  \end{tikzpicture}
\end{subfigure}
\caption{A line search for the maximal shadow. The shadow ``grows'' and
  ``shrinks'' until it contacts the green space visited by the robot.} 
\label{fig:linesearch}
\end{figure}

For a single obstacle with index $i$, we want to find the smallest
$\epsilon_i$ risk bound, or equivalently, largest shadow that contains
the estimate but not the volume of space that the robot may
visit. That is, we wish to solve the following optimization problem:  
\begin{equation*}
\begin{aligned}
& \underset{\epsilon \in (0,1)}{\text{minimize}}
&  \epsilon\\
& \text{subject to}
& &\mathrm{shadow}(\epsilon) \cap X = \varnothing 
\end{aligned}
\end{equation*}
If we restrict ourselves to the shadows obtained by lemma
\ref{lemma:myshadow}, a shadow with a larger $\epsilon$ is strictly
contained in a shadow with a smaller $\epsilon$. This implies that the
intersection is monotone in $\epsilon$, allowing us to solve the above
problem with a line search as shown in
figure~\ref{fig:linesearch}. While we restrict our attention to the
general case, in certain cases, such as where $X$ is a collection of
points, this optimization can be solved analytically.

Essentially we are growing the size of the shadow until it almost touches the states that the robot can visit, $X$. 

% The psuedocode of the algorithm is presented in algorithm \ref{alg:geolinesearch}. 
% \begin{algorithm}
%  \caption{FIND\_MAXIMAL\_SHADOW}
%  \begin{algorithmic}[1]
%  \label{alg:geolinesearch}
%  \renewcommand{\algorithmicrequire}{\textbf{Input:}}
%  \renewcommand{\algorithmicensure}{\textbf{Output:}}
%  \Require $\epsilon_p, \bm \mu, \bm \Sigma$
%  \Ensure  $\epsilon$, s.t. the path is at least $\epsilon$ safe and $\epsilon$ is less than $\epsilon_p$ from minimal $\epsilon$ for which this class of bound may be obtained. 
% % \\ \textit{Initialization} :
%   \State $\epsilon_{low} = 0$
%   \State $\epsilon_{high} = 1$
%   \While {$\epsilon_{high} - \epsilon_{low} \ge \epsilon_p$}
%   \State $\epsilon' = \frac{\epsilon_{high} + \epsilon_{low}}{2}$
%   \State new\_shadow = \Call{generate\_shadow}{$\epsilon', \bm \mu, \bm \Sigma$}
%   \If {\Call{intersects}{new\_shadow}}
%   \State $\epsilon_{high} = \epsilon'$
%   \Else
%   \State $\epsilon_{low} = \epsilon'$
%   \EndIf
%   \EndWhile\\
%  \Return $\epsilon_{low}$ 
%  \end{algorithmic} 
%  \end{algorithm}
 
We define {\sc FIND\_MAXIMAL\_SHADOW}($\epsilon_p, \bm
\mu_i, \bm \Sigma_i, V$), which takes the precision $\epsilon_p$, PGDF
parameters $\bm \mu_i, \bm \Sigma_i$, and swept volume $V$, and
uses a standard bisection search to find and return the largest
$\epsilon$ for which the shadows are non-intersecting with $V$.
This requires $O( \log 1/{\epsilon_p})$ calls of
intersection---proportional to the number of 
digits of precision required. The runtime grows very slowly as the
acceptable probability of collision goes to zero.   
 
\subsection{Multiple Obstacles}
In order to extend the algorithm to multiple obstacles we imitate the
union bound in theorem \ref{lemma:multiobstacle}. We run the line search
to determine the largest allowable $\epsilon$ for every obstacle, and
sum the resulting $\epsilon$'s to get the ultimate bound on the
risk. The psuedocode is presented in algorithm \ref{alg:multiobj}. 

\begin{algorithm}
 \caption{FIND\_MAXIMAL\_SHADOW\_SET}
 \begin{algorithmic}[1]
   \renewcommand{\algorithmicrequire}{\textbf{Input:}}
   \renewcommand{\algorithmicensure}{\textbf{Output:}} \Require
   $\epsilon_p, \bm \mu_i, \bm \Sigma_i, V$ \Ensure $\epsilon$,
   s.t. the path generating volume $V$ is at least $\epsilon$ safe and
   each shadow is less than $\epsilon_p$ away from the minimal
   $\epsilon$ for which this class of bound may be obtained.
% \\ \textit{Initialization} :
  \For {i = 1...n}
  \State $\epsilon_i$ = \Call{find\_maximal\_shadow}{$\epsilon_p, \bm
    \mu_i, \bm \Sigma_i, V$} 
  \EndFor\\
 \Return $\sum \epsilon_i$ 
 \end{algorithmic} 
  \label{alg:multiobj}
 \end{algorithm}
 
This algorithm is embarrassingly parallel because every $\epsilon_i$
can be computed independently without increasing the total amount of
required computation. To obtain a 
total accumulated numerical error less than $\delta$ we only need
to set $\epsilon_p = \delta /n$. If $\omega$ is the complexity of a
single call of intersection, our algorithm runs in $O( \omega n \log n
\log 1/\delta)$ time. However, since the search for shadows can be
done in parallel in a work-efficient manner, the algorithm can run in
$O( \omega \log n \log 1/\delta)$ time on $\Theta(n)$ processors. 

If the intersection check is implemented with a collision checker then finding a safety certificate is only $\log$ factors slower than running a collision check--suggesting that systems robust to uncertainty do not necessarily have to have significantly more computational power. 

Furthermore, since the algorithm computes a separate $\epsilon$ for every obstacle, obstacles with little relevance to the robot's actions do not significantly affect the resulting risk bound. This allows for a much tighter bound than algorithms which allocate the same risk for every obstacle. 
  
\subsection{Experiments}
We can illustrate the advantages of a geometric approach 
by certifying a trajectory with a probability of failure
very close to zero. For an allowable chance of failure of $\epsilon$,
the runtime of sample-based, Monte-Carlo methods tends to depend on $
\frac 1 \epsilon$ as opposed to $\log 1/\epsilon$. Monte-Carlo based
techniques rely on counting failed samples requiring them to run
enough simulations to observe many failed samples. This means that
they have trouble scaling to situations where $\epsilon$ approaches
zero and failed samples are very rare. For example, Janson et al.'s
method takes seconds to evaluate a simple trajectory with $\epsilon =
0.01$, even with variance reduction techniques
\cite{montecarloLJESMP2015}.  

\begin{figure}[tp]
\centering
\includegraphics[width=0.6\columnwidth]{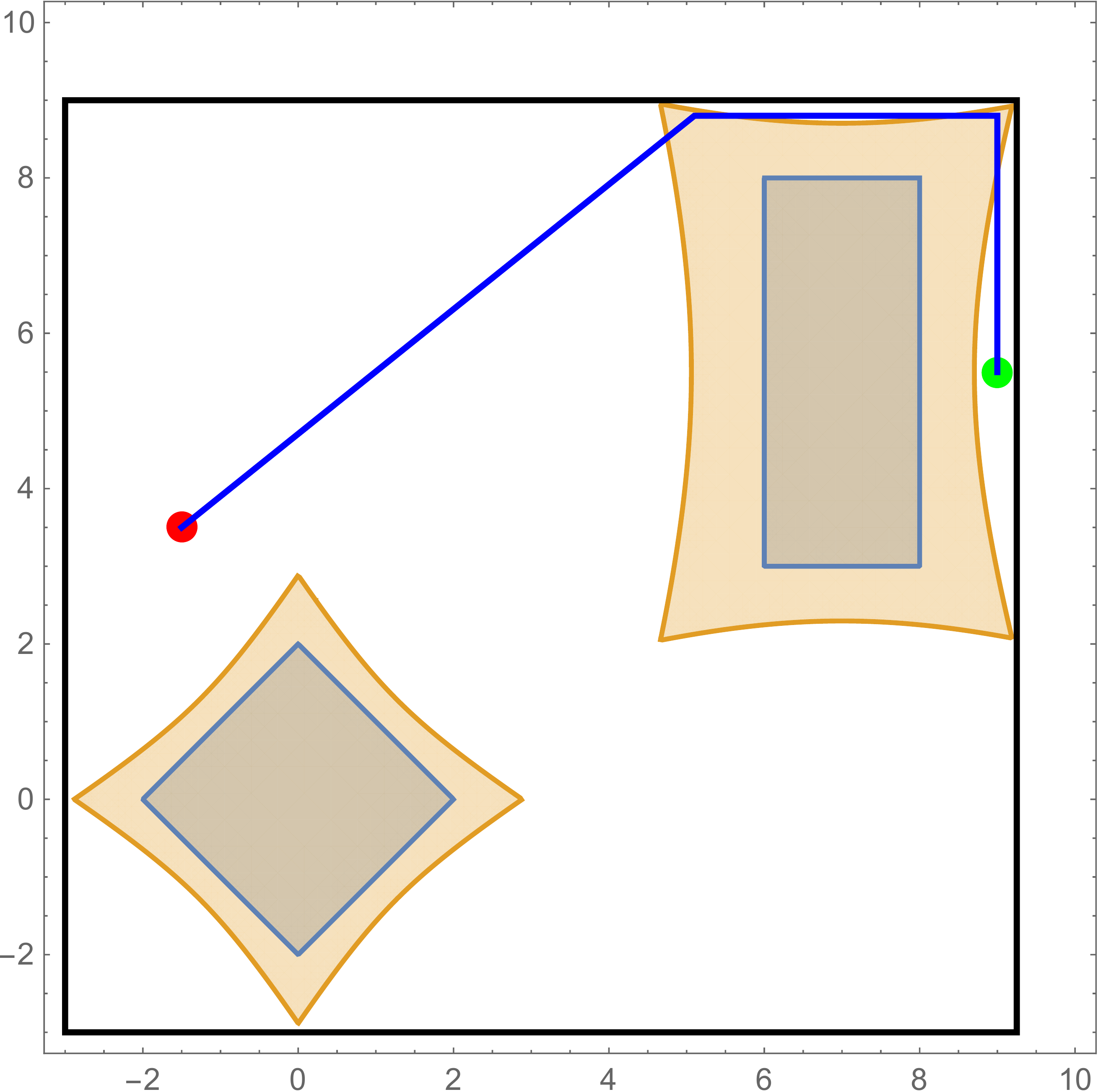}
\caption{Computing the optimal equal allocation of probabilities fails to certify the safety of the path. }
\label{fig:expequalalloc}
\end{figure}

\begin{figure}[tp]
\centering
\includegraphics[width=0.6\columnwidth]{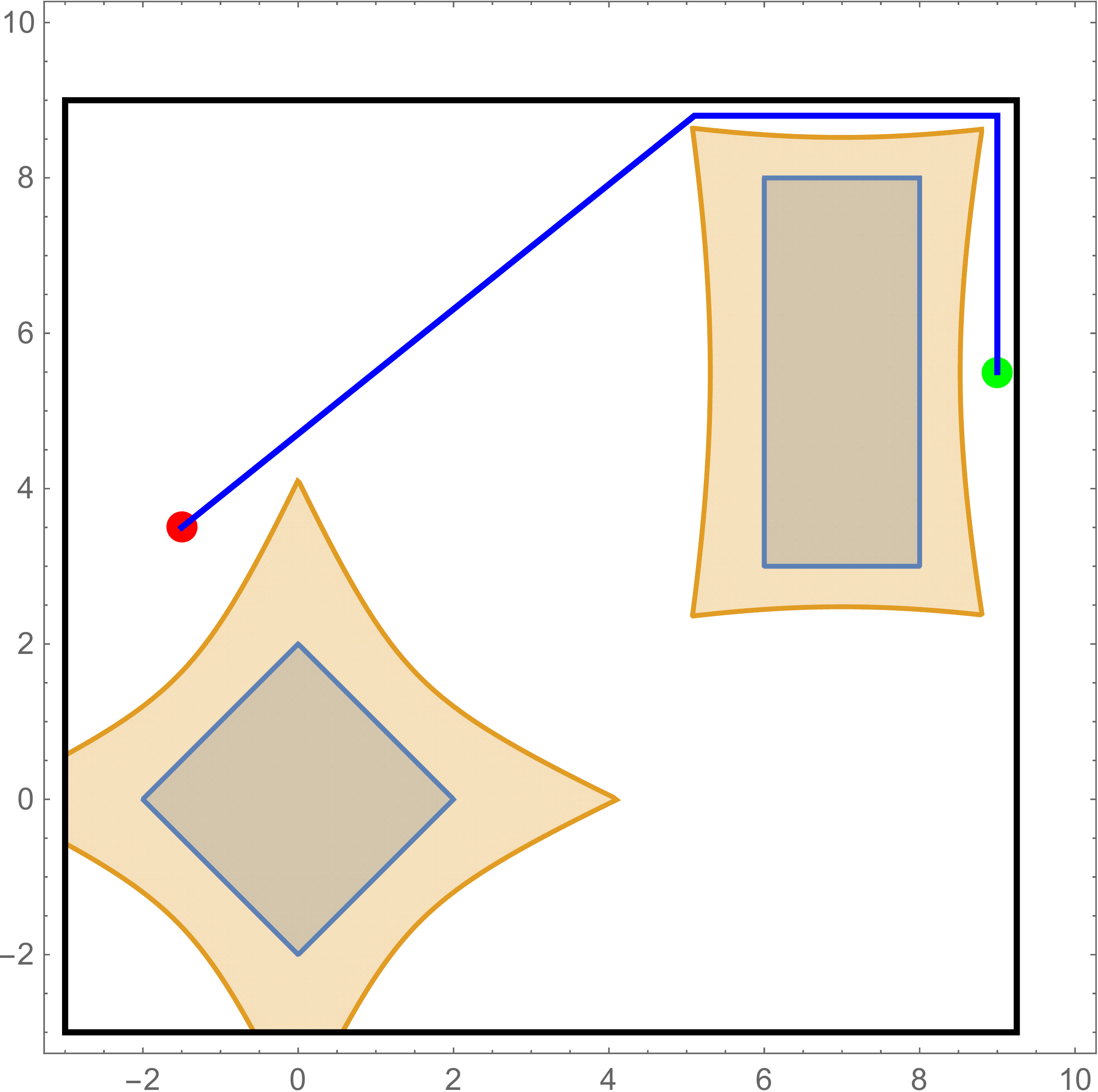}
\caption{Computing the optimal probability for each shadow allows us to successfully verify that the trajectory is safe. }
\label{fig:expoptalloc}
\end{figure}

We demonstrate our algorithm on a simple domain with $\epsilon = 2.2
\times 10^{-5}$. Our algorithm required just 6 calls to a collision
checker for each obstacle. We also demonstrate that our algorithm can
certify trajectories which cannot be certified as safe with shadows of
equal sizes. Figures~\ref{fig:expequalalloc} and \ref{fig:expoptalloc}
show the problem domain. Figure~\ref{fig:expequalalloc} shows that the
trajectory cannot be certified as safe with a uniform risk assigned to
each obstacle. Figure \ref{fig:expoptalloc} shows the shadows found by
our algorithm that prove the trajectory is safe.   

% Note the algorithm is currently all mathematica and requires calls the Brian visual collision checker. 
\section{Online Safety}\label{sec:onlinesafety}
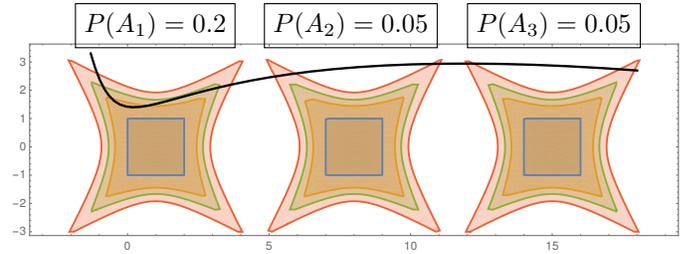
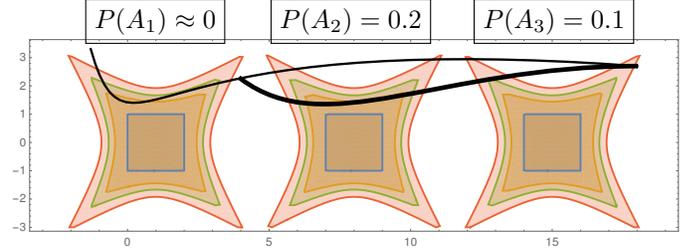
\begin{figure}[tp]
\centering
\begin{subfigure}{\columnwidth}
  \centering
  % old coordinate 
  %     \node[draw,align=left] at (1.9,-0.2) {$P(A) = 0.5$};
  \begin{tikzpicture}
    \node[anchor=south west,inner sep=0] at (0,0) {\includegraphics[width=\columnwidth]{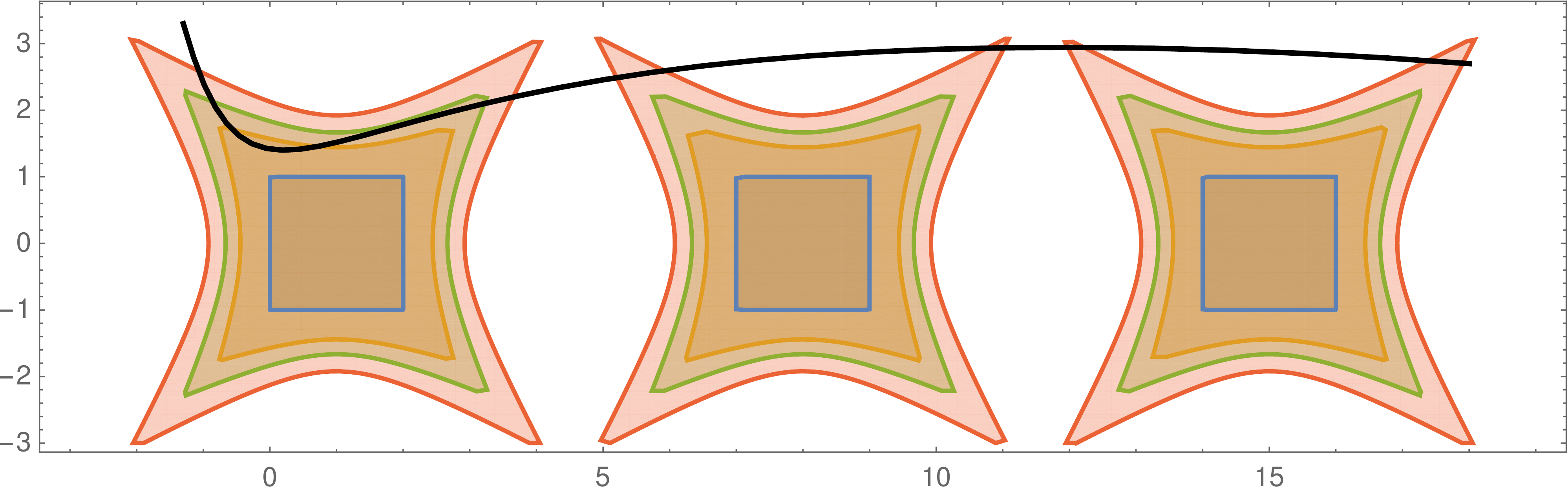}};
    \node[draw,align=left] at (1.9,3) {$P(A_1) = 0.2$};
  \node[draw,align=left] at (4.5,3) {$P(A_2) = 0.05$};
  \node[draw,align=left] at (7.2,3) {$P(A_3) = 0.05$};
  \end{tikzpicture}
  \caption{At first the robot chooses the thin black trajectory which is has a probability of collision of only 0.3. It is more likely to collide with the first obstacle than the remaining obstacles.}
  \label{fig:sub1}
\end{subfigure}%
\\
\begin{subfigure}{\columnwidth}
  \centering
  \begin{tikzpicture}
    \node[anchor=south west,inner sep=0] at (0,0) {\includegraphics[width=\columnwidth]{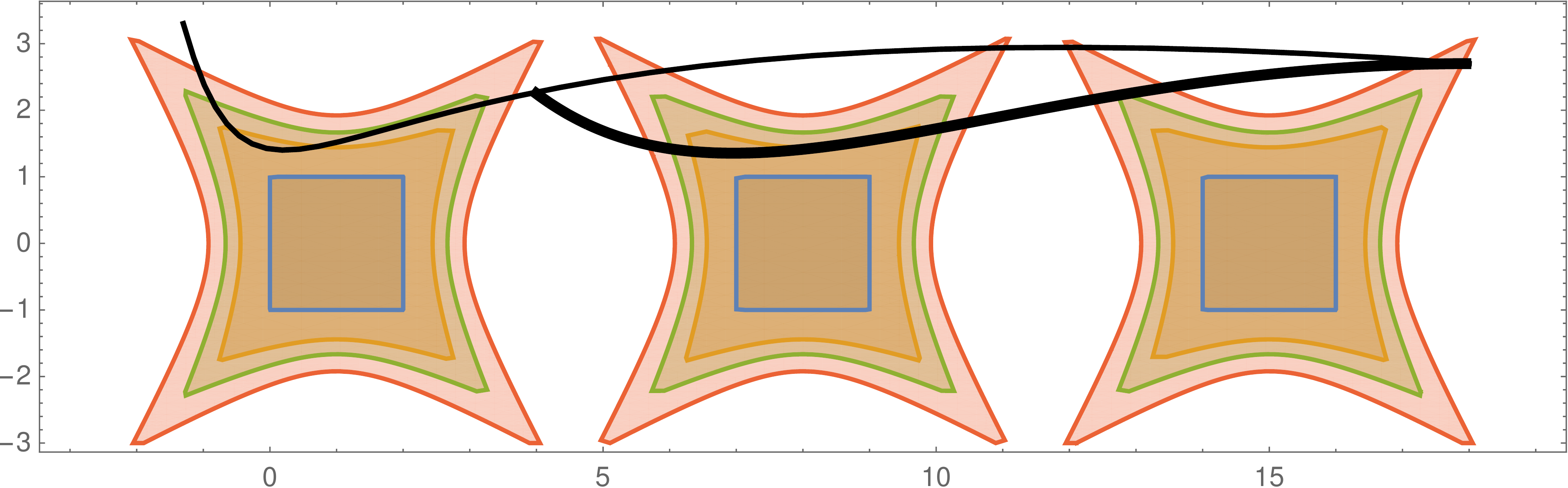}};
    \node[draw,align=left] at (1.9,3) {$P(A_1) \approx 0$};
  \node[draw,align=left] at (4.5,3) {$P(A_2) = 0.2$};
  \node[draw,align=left] at (7.2,3) {$P(A_3) = 0.1$};
  \end{tikzpicture}
  \caption{After observing that it had not collided with the first obstacle, it readjusts the plan to follow the bold trajectory so that the probability of collision is still less than 0.3. However, a system that follows this policy will collide with probability 0.51 even though at every point it was following a trajectory with probability of collision less than 0.3. In other words, for this set of observations $O$, $P( A | \pi, O) > 0.3$. A system that is allowed to change its action after seeing additional information (in this case the fact that it did not collide) must properly account for the risk already taken. }
  \label{fig:sub2}
\end{subfigure}
\caption{Tricking "safety" by changing paths once more information is acquired. Note that even knowing that the system did not collide can serve as information.}
\label{fig:trickingsafety}
\end{figure}

The bounds in the previous section do not immediately generalize to a
setting where the robot acquires more information over time and can be
allowed to change its desired trajectory. Additional care must be
taken to ensure that the system cannot ``trick" the notion of safety
used, and not honor the desired contract on aggregate lifetime risk of
the execution instance. Consider the case where,  if a fair coin turns
up as heads the robot takes a path with a $1.5\epsilon$ probability of
failure and it takes a trajectory with a $0.5 \epsilon$ probability of failure
otherwise. This policy takes an action that is unsafe, but the
probability of failure of the policy is still less than
$\epsilon$. Furthermore, the history of actions is also important in
ensuring aggregate lifetime safety. In figure \ref{fig:trickingsafety}
we illustrate a example of how the robot can always be committed to some
trajectory that is $\epsilon$-safe but have more than an
$\epsilon$ probability of collision over the lifetime of the execution.

Figure \ref{fig:trickingsafety} highlights the need to ensure low
probability of failure under all sets of observations. If this
scenario is run multiple times the failure rate will be much greater
than acceptable. In order to propose an algorithm that allows the
robot to change the desired trajectory as a function of a stream of
information, we develop an alternative criteria for safety that
accounts for risks as they are about to be incurred. We let $p_t$
denote the probability of collision at time $t$, given the information
available at time $t$, given that we follow the trajectory currently
predicted by the policy $\pi$. We note that since the information
itself is random, $p_t$ is a random variable for future times. We say
that a policy $\pi$ is \textit{absolutely safe} if for all times $t$,
equation \eqref{eqn:abssafe} is satisfied. The expectation in the
integral is with respect to the information available at the current
time $t$. 
\begin{align}
\int\limits_{0}^\infty E[p_t \mid \pi ] dt &= \int\limits_0^t p_t dt + \int\limits_t^\infty E[p_t \mid \pi]\;  dt   \le \epsilon \label{eqn:abssafe}
\end{align}     
We note that the $\int\limits_0^t p_tdt$ can be evaluated as an accumulation with standard numerical techniques for evaluating integrals. 

The second term, $\int\limits_t^\infty E[p_t \mid \pi]\; dt$, is exactly the probability that the remaining part of the trajectory will collide and can be evaluated with the method for solving the offline safety problem. 

\subsection{Absolute Safety vs Policy Safety}

% A mathematical characterization of what is necessary to be safe allows
% for the development of a simple safety check that is compatible with
% dynamic policies and replanning as presented in algorithm
% \ref{alg:onlinesafety} 

Algorithm \ref{alg:onlinesafety} provides a method for performing safe
online planning in the case that the PGDF parameters are updated
during execution. While it shows that absolute safety can
be verified efficiently, it is not clear how to efficiently verify
policy safety. However, unlike absolute safety, policy safety
(introduced in section~\ref{sec:policySafety}) is a
very direct condition on aggregate lifetime probability of collision
and can be easier to interpret. In this section we compare policy
safety to absolute safety in order to identify when they are
equivalent.  

First we show that absolute safety is a strictly stronger condition
than policy safety in theorem \ref{thm:absoluteimpliespolicy}. This
comes by integrating the probability of failure over time to get the
total probability of failure. Since the absolute safety condition in
equation \eqref{eqn:abssafe} must always be satisfied, regardless of
the observation set, the probability of failure for that information
set will always be sufficiently small. 

\begin{thm}\label{thm:absoluteimpliespolicy} If a policy is absolutely safe, then it is also safe in the policy safety sense.  
\end{thm}
{\bf Proof:} Please see the supplementary material.

\begin{comment}
\begin{proofatend}
Assume for the sake of contradiction that a policy is absolutely safe, but not policy safe. That means there exists a set of observations $O$ and time $t$ for which policy safety does not hold, but absolute safety does not. 

The probability of collision conditioned on these observations is $\int\limits_0^t p_t | O dt + \int\limits_{t}^\infty E[p_t | O] dt$. Since the integral is less than $\epsilon$ so must the probability of collision. However, this implies that the system is policy safe for this set of observations $O$ and time $t$, yielding a contradiction. 

Thus if a policy is absolutely safe it must also be policy safe.  
\end{proofatend}
\end{comment}

Absolute safety, however, is not always equivalent to policy safety. The key difference lies in how the two conditions allow future information to be used. Absolute safety requires that the system always designate a safe trajectory under the current information while policy safety allows the robot to postpone specifying a complete, safe trajectory if it is certain it will acquire critical information in the future. 

In order to formalize when policy safety and absolute safety are
equivalent, we introduce the notion of an information adversary. An
information adversary is allowed to (1) see the observations at the same
time as the agent, (2) access the policy used by the agent, and (3)
terminate the agent's information stream at any point. Policy safety
under an information adversary is guaranteed by the policy
safety conditions if the information stream can stop naturally at any
point. Theorem \ref{thm:infoadv} shows that policy safety with an
information adversary is equivalent to absolute safety.
 
\begin{thm}\label{thm:infoadv} A policy that is safe at all times
  under an information adversary is also absolutely safe.  
\end{thm}
{\bf Proof:} Please see the supplementary material.

\begin{comment}
\begin{proofatend}
Assume for the sake of contradiction that there exists set of observations $O$ and a time $t$ for which absolute safety does not hold. That is to say that conditioned on these observations 
\begin{align*}
\int\limits_{0}^t p_t dt + \int\limits_t^\infty E[p_t | \pi] dt > \epsilon
\end{align*}
Let the information adversary stop the flow of information to the robot at time $t$. Let $A_1$ denote the event that the system fails during times $(0, t]$ and $A_2$ denote the event that the systems fails during times $(t, \infty)$. We note that a system can fail at most once, so $A_1, A_2$ are exclusive. 

Utilizing the fact that $\int\limits_t^\infty E[p_t| O] dt = P(A_2 | O)$ and $\int\limits_0^t p_t | O dt = P(A_1 | O)$, we get that the probability of failure $P(A_1 \cup A_2) = P(A_1) + P(A_2) \ge \epsilon$ violating our assumption that the system is policy safe under the set of observations $O$.

Thus if a system is policy safe under an information adversary, it is also absolutely safe.  
\end{proofatend}
\end{comment}

\begin{algorithm}[tp]
 \caption{FIND\_MAXIMAL\_SHADOWS\_ONLINE}
 
\label{alg:onlinesafety}
 \begin{algorithmic}[1]
 \renewcommand{\algorithmicrequire}{\textbf{Input:}}
 \renewcommand{\algorithmicensure}{\textbf{Output:}}
 \Require intersects$_i$, $\epsilon_p$, $t$, $\{p_{t'} |  \forall t' \in [0,t] \}$
 \Ensure  $\epsilon$, s.t. $\epsilon$ is greater than the sum cumulate risk taken before the current time $t$ and the future risk. $\epsilon$ is at most $\epsilon_p$ away from the minimal $\epsilon$ for which a bound of this class may be obtained. 
% \\ \textit{Initialization} :
\State $\epsilon_1  = \int\limits_0^t p_t dt$
\State $\epsilon_2 =$ \Call{find\_maximal\_shadow\_set}{$\epsilon_p, \bm \mu_{1...n}, \bm \Sigma_{1...n}$} \\
 \Return $\epsilon_1 + \epsilon_2$ 
 \end{algorithmic} 
 \end{algorithm}
 
\subsection{Experiments}
\begin{figure}[tp]
\centering
\includegraphics[width=\columnwidth]{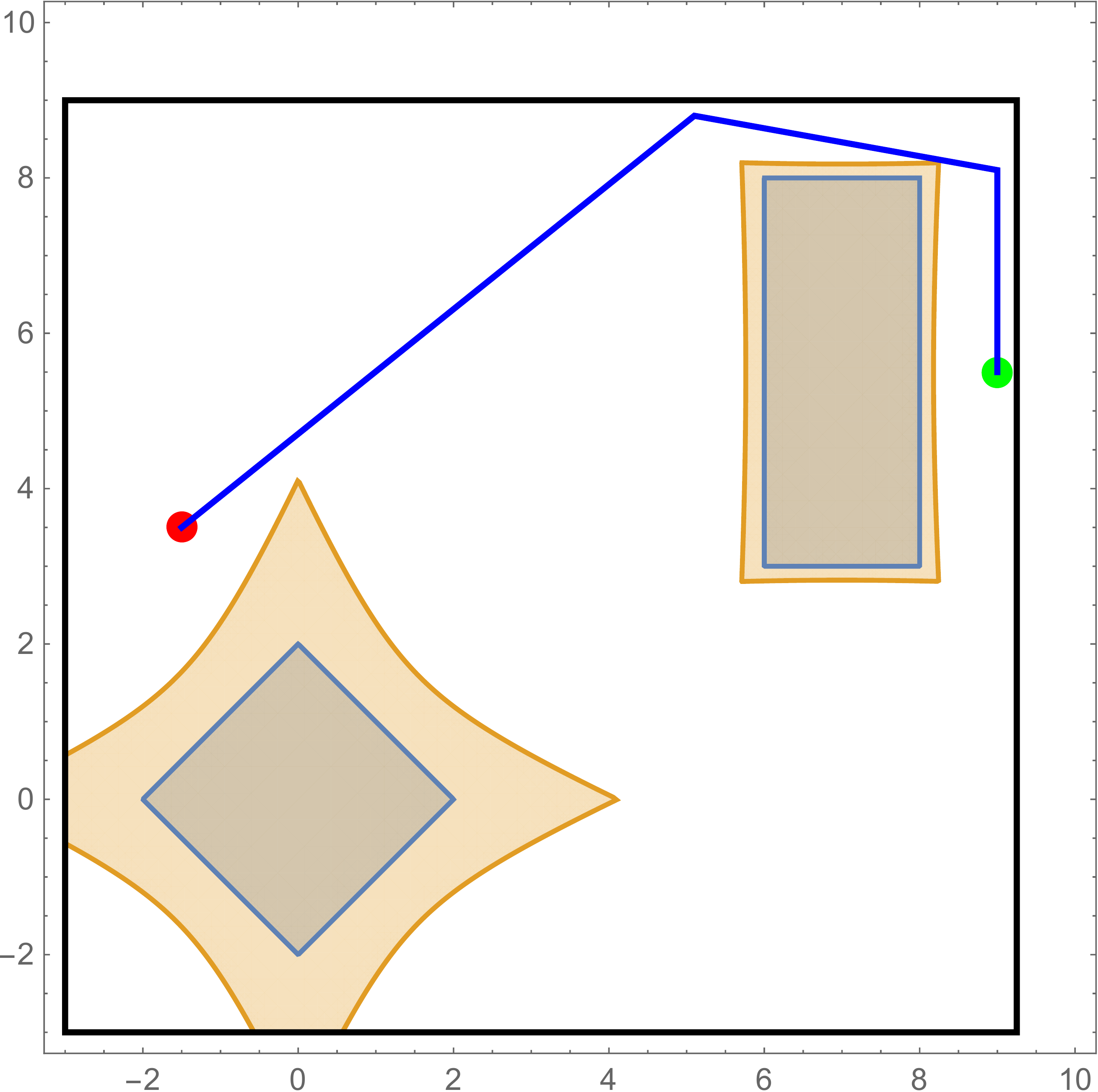}
\caption{Once the robot reached the top of the old trajectory it got a new observation regarding the second obstacle. This allows it to shrink the shadow around the second obstacle and take the less conservative path that is shown above.}
\label{fig:replanexample}
\end{figure}

We demonstrate a simple replanning example based on the domain
presented in figure \ref{fig:expoptalloc}. Once the robot gets halfway
through,  it will receive a new observation that helps it refine its
estimate of the larger, second obstacle. This allows it to shrink the
volume of the shadow corresponding to the same probability and certify
a new, shorter path as safe. This new path is shown in figure
\ref{fig:replanexample}. It takes this new trajectory without taking
an unacceptable amount of risk.  

\section{Enabling Safe Planning using Safety Certificates}
The safety certification algorithms we presented above can be used for more than just checking safety. It can enable safe planning as well. We present a modification to the RRT algorithm that restricts output to only safe plans \cite{lavalle1998rapidly}. Every time the tree is about to be expanded, the risk of the trajectory to the node is computed. The tree is only grown if the risk of the resulting trajectory is acceptable.

\begin{algorithm}[tp]
\caption{SAFE\_RRT}
 
\label{alg:saferrt}
 \begin{algorithmic}[1]
 \renewcommand{\algorithmicrequire}{\textbf{Input:}}
 \renewcommand{\algorithmicensure}{\textbf{Output:}}
 \Require $\epsilon_{safe}$, $\epsilon_p,t, q_s,q_f, \bm \mu_{1...n}, \bm \Sigma_{1...n}$
 \Ensure  A sequence of waypoints from $q_s$ to $q_f$, such that the trajectory going through these waypoints has a probability of less than $\epsilon_{safety}$ of collision.  
% \\ \textit{Initialization} :
\State $tree = $ new \Call{tree}{$q_s$}
\For {$\mathrm{iteration} = 1...n$}
\State $x_{rand} =$ \Call{random\_state}{} 
\State $x_{near} =$ \Call{nearest\_neighbor}{TREE}
\State $x_{new} =$ \Call{extend}{$x_{near}, x_{rand}$}
\State $X =$ \Call{get\_trajectory}{$tree, x_{near}, x_{new}$}
\State $risk =$\\ \hspace{0.9cm}\Call{find\_maximal\_shadow\_set}{$X, \epsilon_p, \bm \mu_{1...n}, \bm \Sigma_{1...n}$} % \mathrm{find\_maximal\_safe\_epsilon}(traj)$
\If {$ risk \le \epsilon_{safe}$}
\State \Call{add\_child}{$tree, x_{near}, x_{new}$} 
\EndIf
\EndFor \\
\Return $tree$
 \end{algorithmic} 
 \end{algorithm}
 
We note that it is not necessary to check the safety of the entire
trajectory every time the tree is extended. Since the bounds for each
obstacle are determined by a single point in the trajectory, it is
sufficient to consider the following two risks: the trajectory from the
root of the tree to $x_{near}$ and the trajectory from $x_{near}$ to
$x_{new}$. Finally we note that analyzing probabilistic completeness
is quite different in the risk constrained case from the original
case. We do not believe this method is probabilistically
complete. Unlike the deterministic planning problem, the trajectory
taken to reach a point affects the ability to reach future
states---breaking down a crucial assumption required for RRTs to be
probabilistically complete.

\begin{figure}[tp]
\centering
\includegraphics[width=\columnwidth]{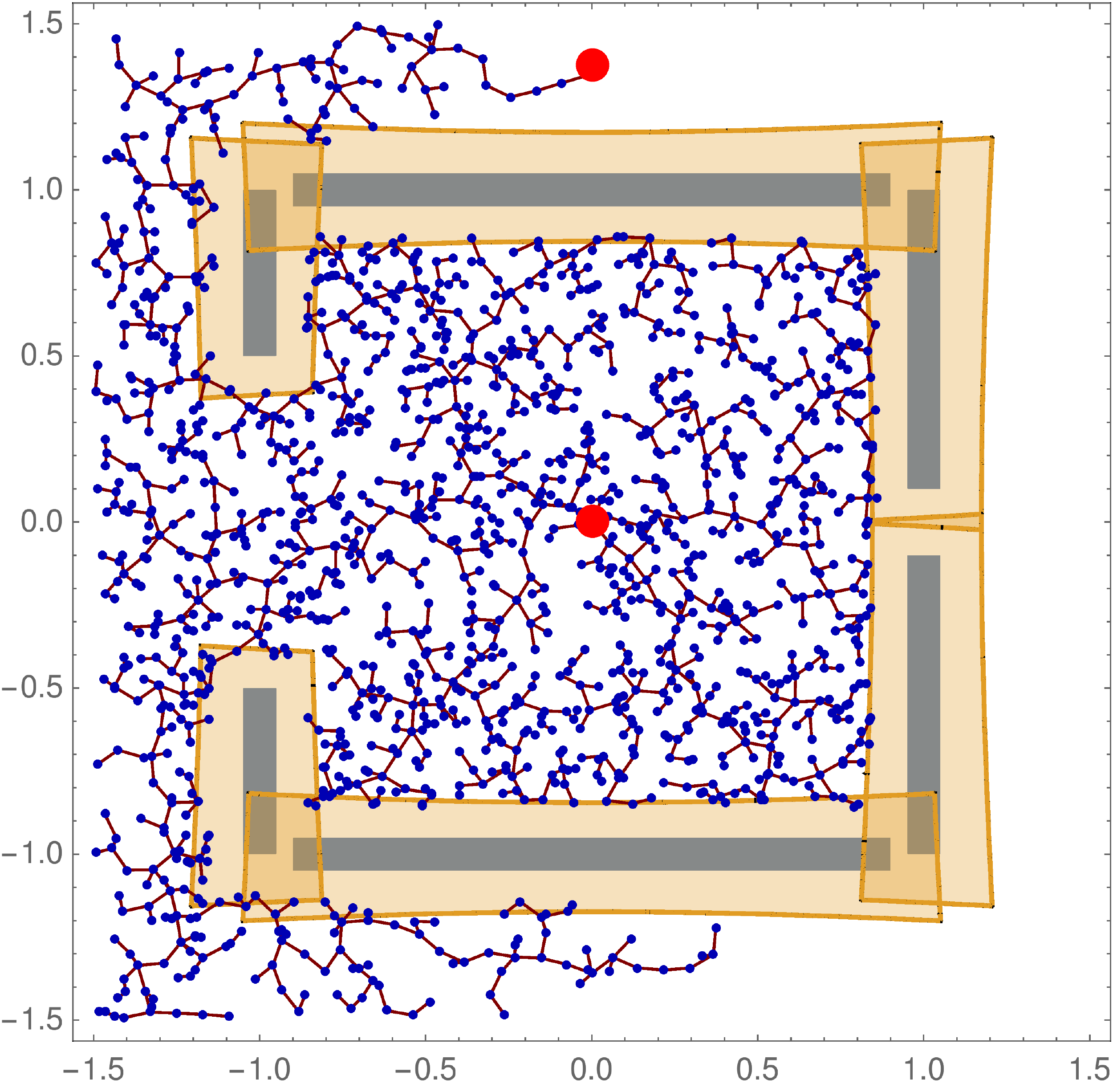}
\caption{A tree of safe trajectories branching from the red dot in interior of the box. Equally sized shadows are shown for reference.}
\label{fig:safeplan}
\end{figure}

We demonstrate the safe-RRT algorithm on a point robot trying to escape a box. The box has two exits. While the robot can safely pass through the larger exit, it cannot safely pass through the smaller exit. The planner is run to only return paths with a probability of failure less than $0.5\%$. Figure \ref{fig:safeplan} shows a safe tree from the red dot inside the box to the red dot above the box. Figure \ref{fig:safetraj} shows just the ultimate trajectory with its corresponding shadows that certify the probability of failure as less than $0.26\%$. Note that some shadows did not extend all the way to the trajectory as their risk was already below the numerical threshold. 

The experiment shown in figure \ref{fig:safeplan} demonstrates offline-safety. If the robot were given additional information during execution, we could use the equations for online-safety to re-run the RRT with the new estimates of obstacles while preserving the safety guarantee. 

\section{Conclusion}

We presented a framework to compute shadows, the geometric equivalent
of a confidence interval, around observed geometric objects. Our bounds are
tighter than those of previous methods and, crucially, the tightness of the
bounds does not depend on the number of obstacles. In order to achieve
this tightness we rely on computing a bound specific to a trajectory
instead of trying to identify a generic ``safe" set.  

We present offline and online variants of algorithms that can verify safety with respect to the shadows identified above for both trajectories and policies. The online method highlights nuances and potential issues with a mathematical definition of safety, and presents a strong, but still computationally verifiable notion of safety. These algorithms do not have a computational complexity much larger than a collision check, and are only a $O\left (\log n \log \frac 1 \epsilon \right )$ factor slower than a collision check for $n$ obstacles and an $\epsilon-$safety guarantee. Finally the output of these algorithms is easy to verify, allowing the output to serve as a safety certificate.

These safety certification algorithms are an important not only in ensuring that a given action is safe, but also in enabling the search for safe plans. We demonstrate an extension to the RRT algorithm that only outputs safe plans. 

Future work includes using other models of random geometry. A model for occupancy would match the information of interest to a motion planner more closely, and may be easier to apply to real world data sets. The presented algorithms are compatible with mixture models---a potentially interesting and useful direction that should be evaluated empirically. Another direction of interest is the development of safe planning algorithms with strong theoretical guarantees such as optimality and probabilistic completeness. 

\section*{Acknowledgements}
BA would like to thank Gireeja Ranade, Debadeepta Dey and Ashish Kapoor for introducing him to the safe robot navigation problem during his time at Microsoft Research. 

We gratefully acknowledge support from the Thomas and Stacey Siebel Foundation, from NSF grants 1420927 and 1523767, from ONR grant N00014-14-1-0486, and from ARO grant W911NF1410433. Any opinions, findings, and conclusions or recommendations expressed in this material are those of the authors and do not necessarily reflect the views of our sponsors.

%\section*{Proofs}
%\printproofs

%% Use plainnat to work nicely with natbib. 

\bibliographystyle{plainnat}
\bibliography{references}
\pagebreak 
\setcounter{thm}{0}
\setcounter{lemma}{0}

\section*{Proofs}
 \begin{thm}\label{lemma:lgshape} Consider $\epsilon \in (0,1)$,
   $\bm n \sim \mathcal N \left  (\bm \mu, \bm \Sigma \right)$ such
   that the combination of $\epsilon, \bm \mu, \bm \Sigma$ is
   non-degenerate. There exists $X$ s.t. $\bm n^T \bm x \ge 0$ is
   contained within $X$ with probability at least $1 -
   \epsilon$. \end{thm}

\begin{proof} \textit{Preliminaries:} In order to avoid a constant term in the definition of halfspaces we use homogeneous coordinates, and thus our shadow will be defined asinvariant under positive scaling. $x$ will always refer to these coordinates. Likewise, parameterizations of hyperplanes that go through the origin are postive scale invariant $\alpha$ will always refer to a hyperplane parametrization.

It is important to note that $\alpha$'s act on the $x$'s. They belong to different spaces, even though they may share geometric properties and dimensionality.

We will for the convenience of applying standard theorems from geometry, sometimes treat these spaces by looking at the representation as a subset of $\mathbb R^{n}$, where the sets will be closed under positive scaling. This will also prove convenient when working with the gaussian distribution, as it is not typically defined on such spaces.

After the proof, we will provide an interpretation of the theorem in standard coordinates ($\mathbb R^{n-1}$) in order to provide intuition for the shape of the generated shadows.

\textit{Proof Sketch:}
\begin{enumerate}
\item Identify a set of parameters, as a scaled covariance ellipsoid around the mean, that has probability mass of $1 - \epsilon$.
\item Consider the set of hyperplanes they form. This will be a linear cone.
\item Take any $x \in \mathbb R^{n}$ that is contained in by any $\alpha$ in the previous set of hyperplanes. This will be the shadow. Its complement will also be a linear cone.
\item Converting back to non-homogeneous coordinates will reveal that the shape of the complement of a shadow is a conic section.
\end{enumerate}

First we note that the squared radius values of the $n-d$ standard multivariate Gaussian distribution is distributed as a chi-squared distribution with $n$ degrees of freedom. Let $\phi(k)$ be the corresponding cumulative distribution function. Then it follows that the probability mass of $A = \{ \bm \alpha | (\bm \alpha - \bm \mu)^T \bm \Sigma^{-1} (\bm \alpha - \bm \mu \le \phi^{-1}(1 - \epsilon)\}$ is $1 - \epsilon$.

We note, however, that since $\alpha$'s are invariant under positive scaling, the resulting hyperplanes form a linear cone, $C$. This is the minimal cone with axis $\bm \mu$ that contains the above ellipsoid.

We consider the polar of this cone, $C^0$. A polar cone $C^0$ of $C$ is $\{ x | x^T y \le 0 \forall y \in C\}$. Since every element of $C^0$ has a positive negative dot product with every element of $C$, ${C^0}^C$ is the set of points which have a positive dot product with at least one element of $C$. ${C^0}^C$ is the set that is contained by at least a single $\alpha \in A$ and an $\epsilon-$shadow. We note that $C^0$, the complement of the shadow, is also a linear cone since the polar cone of a linear cone is also a linear cone.

In order to understand the geometry this induces on non-homogeneous coordinates we take a slice where the last coordinate is equal to $1$. Since our set has an equivalence class of positive scaling, this slice will contain exactly one representative of every point. This set is a standard conic section---giving us the familiar curves we see throughout the paper.

If this slice has finite (ex. 0) measure in $\mathbb R^{n-1}$ we say that the shadow is degenerate---it did not provide us with a useful shadow. One example of a degenerate case is when the set $A$ contained the origin. A neighborhood of the origin in parameter space produces halfspaces that contain all of $\mathbb R^n$. The distribution being close to centered about the origin represents having little to no information about the points contained within the half-space.
\end{proof}

\begin{lemma}\label{lemma:polybuild} Consider an polytope defined by $\bigcap\limits_i \bm n^T \bm x \le 0$. Let $X_i$ be a set that contains the halfspace defined by $\bm n_i$ with probability at least $ 1- \epsilon_i$ (for example as in theorem \ref{lemma:lgshape}). Then $\bigcap\limits_i X_i$ contains the polytope with probability at least $1 - \sum\limits_i \epsilon_i$.
\end{lemma}
\begin{proof}
    We note that $\bigcap\limits_i X_i$ contains the polytope if every $X_i$ contains its corresponding halfspace. Since the probability that $X_i$ does not contain its corresponding halfspace is $\epsilon_i$, a union bound gives us that the probability that any $X_i$ does not contain its corresponding halfspace is bounded by $\sum\limits_i \epsilon_i$. $\bigcap\limits_i X_i$ containing the polytope is the complement of this event, thus the probability that $\bigcap\limits_i X_i$ contains the polytope is at least $ 1 - \sum\limits_i \epsilon_o$
\end{proof}

\begin{lemma}\label{lemma:myshadow} If an obstacle is PGDF with
      nondegenerate parameters and $m$ sides, we can construct an
        $\epsilon$-shadow as the intersection of the $\epsilon/m$ shadows of
          each of its sides.
      \end{lemma}
\begin{proof}
Let $X_i$ be the shadow constructed by Lemma 1 with parameter
$\epsilon/m$ for side $i$. Then the shadow $X = \bigcap\limits_i X_i$
contains the PGDF with probability at least $1 -
\sum\limits_i\epsilon_i/m = 1 - \epsilon$.
\end{proof}

\begin{thm}\label{lemma:multiobstacle} Let $X$ be the set of states that the system may visit during its lifetime. If for a given set of obstacles, indexed by $i$, and their corresponding $\epsilon_i$ shadows, $X$ does not intersect any shadow, then the probability of collision with any obstacle over the lifetime of the system is at most $\sum \epsilon_i$.
\end{thm}
\begin{proof}
    Let $A_i$ be the event that the intersection of $X$ and obstacle $i$ is nonempty. Since the shado
    w of obstacle $i$ did not intersect $X$, this means that obstacle $i$ must not be contained by it
    s shadow. The probability of this is less than $\epsilon_i$ by definition of $\epsilon-$shadow. T
    hus the probability of $A_i$ must be at most $\epsilon_i$.

    Applying a union bound gives us that $P(\bigcup A_i) \le \sum \epsilon_i$.
\end{proof}

\begin{lemma} \label{lemma:uniontight} 
Given $n$ obstacles and their shadows, if
\begin{itemize}
\item the events that each obstacle is not contained in its
shadow are independent,
\item the probability that obstacles are
not contained in their shadows is less than $\epsilon$, and
\item $\epsilon = O\left (\frac{\sqrt \delta}{n}\right )$
\end{itemize}
then the difference between the true
probability of a shadow not containing the object, and the union
bound in theorem~\ref{lemma:multiobstacle} is less than $\delta$.
\end{lemma}  % TODO word better
\begin{proof}
We note that the union bound is only loose when at least two events happen at the same time, so it is sufficient to bound the probability that two shadows fail to contain their corresponding PGDF.

The probability of this happening for two given obstacles is at most $\epsilon^2$, and there are $n \choose 2$ such combinations. Thus by a union bound, the probability of two shadows failing to contain their obstacle is at most ${n \choose 2} \epsilon^2$. If $\epsilon \le k \frac{\sqrt \delta}{n}$ for $k \ge \frac{1}{\sqrt 2}$, then the probability of this happening is at most $\delta$.

Since the probability of more than one event happening at the same time is bounded by $\delta$, the probability given by the union bound is within $\delta$ of the true probability.
\end{proof}

\begin{thm}\label{thm:absoluteimpliespolicy} If a policy is absolutely safe, then it is also safe in the policy safety sense.
\end{thm}
\begin{proof}
Assume for the sake of contradiction that a policy is absolutely safe, but not policy safe. That means there exists a set of observations $O$ and time $t$ for which policy safety does not hold, but absolute safety does not.

The probability of collision conditioned on these observations is $\int\limits_0^t p_t | O dt + \int\limits_{t}^\infty E[p_t | O] dt$. Since the integral is less than $\epsilon$ so must the probability of collision. However, this implies that the system is policy safe for this set of observations $O$ and time $t$, yielding a contradiction.

Thus if a policy is absolutely safe it must also be policy safe.
\end{proof}

\begin{thm}\label{thm:infoadv} A policy that is safe at all times under an information adversar
y is also absolutely safe.
\end{thm}
\begin{proof}
Assume for the sake of contradiction that there exists set of observations $O$ and a time $t$ for
 which absolute safety does not hold. That is to say that conditioned on these observations
\begin{align*}
\int\limits_{0}^t p_t dt + \int\limits_t^\infty E[p_t | \pi] dt > \epsilon
\end{align*}
Let the information adversary stop the flow of information to the robot at time $t$. Let $A_1$ de
note the event that the system fails during times $(0, t]$ and $A_2$ denote the event that the sy
stems fails during times $(t, \infty)$. We note that a system can fail at most once, so $A_1, A_2
$ are exclusive.

Utilizing the fact that $\int\limits_t^\infty E[p_t| O] dt = P(A_2 | O)$ and $\int\limits_0^t p_t
 | O dt = P(A_1 | O)$, we get that the probability of failure $P(A_1 \cup A_2) = P(A_1) + P(A_2)
\ge \epsilon$ violating our assumption that the system is policy safe under the set of observatio
ns $O$.

Thus if a system is policy safe under an information adversary, it is also absolutely safe.
\end{proof}

\end{document}